\documentclass[10pt,journal,compsoc]{IEEEtran}
\ifCLASSOPTIONcompsoc
  \usepackage[nocompress]{cite}
\else
  \usepackage{cite}
\fi
\ifCLASSINFOpdf
  \usepackage[pdftex]{graphicx}
\else
\fi
\usepackage{amsmath}
\usepackage{amsfonts}
\usepackage{amsmath}
\usepackage{multirow}
\usepackage{multicol}
\usepackage{subcaption}
\usepackage{array}
\usepackage{booktabs}
\usepackage{makecell}
\usepackage{bm}
\usepackage{adjustbox}
\usepackage[dvipsnames]{xcolor}

\usepackage{amssymb}
\usepackage{amsthm}
\theoremstyle{definition}
\newtheorem{definition}{Definition}
\theoremstyle{theorem}
\newtheorem{theorem}{Theorem}[subsection]
\newtheorem{corollary}[theorem]{Corollary}
\theoremstyle{definition}
\newtheorem{example}{Example}
\theoremstyle{remark}
\newtheorem{remark}{Remark}

\usepackage{pgfplots}
\usepackage{tikz}

\definecolor{bblue}{HTML}{FFE333}
\definecolor{rred}{HTML}{7FFF00}
\definecolor{ggreen}{HTML}{87CEEB}

\usepackage{algorithm}
\usepackage{algorithmic}
\usepackage{tabularx}
\usepackage{lipsum}
\usepackage{soul}
\newcolumntype{Y}{>{\centering\arraybackslash}X}

\begin{document}
\usepgfplotslibrary{groupplots}
%
\title{PAGE: Prototype-Based Model-Level \\Explanations for Graph Neural Networks}

\author{Yong-Min Shin, {\em Student Member}, {\em IEEE}, Sun-Woo Kim, \\ and Won-Yong Shin, {\em Senior Member}, {\em IEEE}
\IEEEcompsocitemizethanks{
\IEEEcompsocthanksitem Y.-M. Shin is with the School of Mathematics and Computing (Computational Science and Engineering), Yonsei University, Seoul 03722, Republic of Korea. E-mail: jordan3414@yonsei.ac.kr.
\IEEEcompsocthanksitem S.-W. Kim was with the Department of Statistics, Yonsei University, Seoul 03722, Republic of Korea. He is now with the Kim Jaechul Graduate School of AI at Korea Advanced Institute of Science and Technology (KAIST), Seoul 02455, Republic of Korea. E-mail: kswoo97@kaist.ac.kr.
\IEEEcompsocthanksitem W.-Y. Shin is with the School of Mathematics and Computing (Computational Science and Engineering), Yonsei University, Seoul 03722, Republic of Korea, and also with the Graduate School of Artificial Intelligence, Pohang University of Science and Technology (POSTECH), Pohang 37673, Republic of Korea. E-mail: wy.shin@yonsei.ac.kr.\protect\\
(Corresponding author: Won-Yong Shin.)}}
%
%


\markboth{}%
{Shin \MakeLowercase{\textit{et al.}}: PAGE: Prototype-based model-level explanations for graph neural networks}

\IEEEtitleabstractindextext{%
\begin{abstract}

Aside from graph neural networks (GNNs) \textcolor{black}{attracting} significant attention as a powerful framework revolutionizing graph representation learning, there has been an increasing demand for explaining GNN models. Although various explanation methods for GNNs have been developed, most studies have focused on {\em instance-level} explanations, which produce explanations tailored to a given graph instance. In our study, we propose {\em Prototype-bAsed GNN-Explainer (\textsf{PAGE})}, a novel {\em model-level} GNN explanation method that explains what the underlying GNN model has learned for graph classification by discovering human-interpretable {\em prototype graphs}. Our method produces explanations for a given {\em class}, thus being capable of offering more concise and comprehensive explanations than those of instance-level explanations. First, \textsf{PAGE} selects embeddings of class-discriminative input graphs on the graph-level embedding space after clustering them. Then, \textsf{PAGE} discovers a common subgraph pattern by iteratively searching for high matching node tuples using node-level embeddings via a {\em prototype scoring} function, thereby yielding a prototype graph as our explanation. Using six graph classification datasets, we demonstrate that \textsf{PAGE} qualitatively and quantitatively outperforms the state-of-the-art model-level explanation method. We also carry out \textcolor{black}{systematic experimental studies} by \textcolor{black}{demonstrating} the relationship between \textsf{PAGE} and instance-level explanation methods, the robustness of \textsf{PAGE} to input data scarce environments, and the computational efficiency of the proposed prototype scoring function in \textsf{PAGE}.
\end{abstract}

}

\maketitle

\IEEEdisplaynontitleabstractindextext

\IEEEpeerreviewmaketitle

\IEEEraisesectionheading{\section{Introduction}\label{sec:introduction}}
\subsection{Background and Motivation}
\IEEEPARstart{G}{raphs} are a ubiquitous way to organize a diverse set of complex real-world data such as social networks and molecular structures. Graph neural networks (GNNs) have been widely studied as a powerful means to extract useful features from underlying graphs while performing various downstream graph-related tasks~\cite{wu2019gnnsurvey}. GNNs are known to have high expressive capability via message passing \textcolor{black}{to} effectively \textcolor{black}{learn} representations of both nodes and graphs~\cite{xu2019gin}, while adopting neural networks as a basic building block to learn the graph structure and node features (attributes).

Despites their strengths, such GNN models lack transparency since they do not offer any human-interpretable explanations of GNN's predictions for a variety of downstream tasks. Thus, fostering {\it explainability} for GNN models has become of recent interest as it enables a thorough understanding of the model's behavior as well as trust and transparency~\cite{ribeiro2016lime}. Recent attempts to explain GNN models mostly highlight a subgraph structure within a given input graph that contributed most towards the underlying GNN model's prediction~\cite{baldassarre2019infectionsolubility, ying2019gnnexplainer, yuan2021subgraphx, luo2020pgexplainer, schlichtkrull2021graphmask, schnake2021gnnlrp, vu2020pgmexplainer, lucic2022cfgnnexplainer, wang2022rcexplainer}. These so-called {\it instance-level} explanation methods for GNNs provide an in-depth analysis \textcolor{black}{for a} given graph instance~\cite{yuan2020xaisurvey}.

Nevertheless, instance-level GNN explanations do not reveal the {\em general} behavior of the underlying model that has already been trained over \textcolor{black}{an entire} dataset, consisting of numerous graphs. In some real-world scenarios, explanations that showcase the decision-making process of GNN models are indeed beneficial. For example, when a black-box GNN model is employed in \textcolor{black}{a} real-world setting, it is likely that the model will encounter various instances \textcolor{black}{unobserved} during training. In such a case, understanding \textcolor{black}{the} general decision-making process of the GNN model will help \textcolor{black}{predict} the GNN's behavior in new environments. This motivates us to study {\it model-level} explanations that aim to interpret GNNs at the model-level.

\begin{figure}[t]
\centering
\includegraphics[width=0.9\columnwidth]{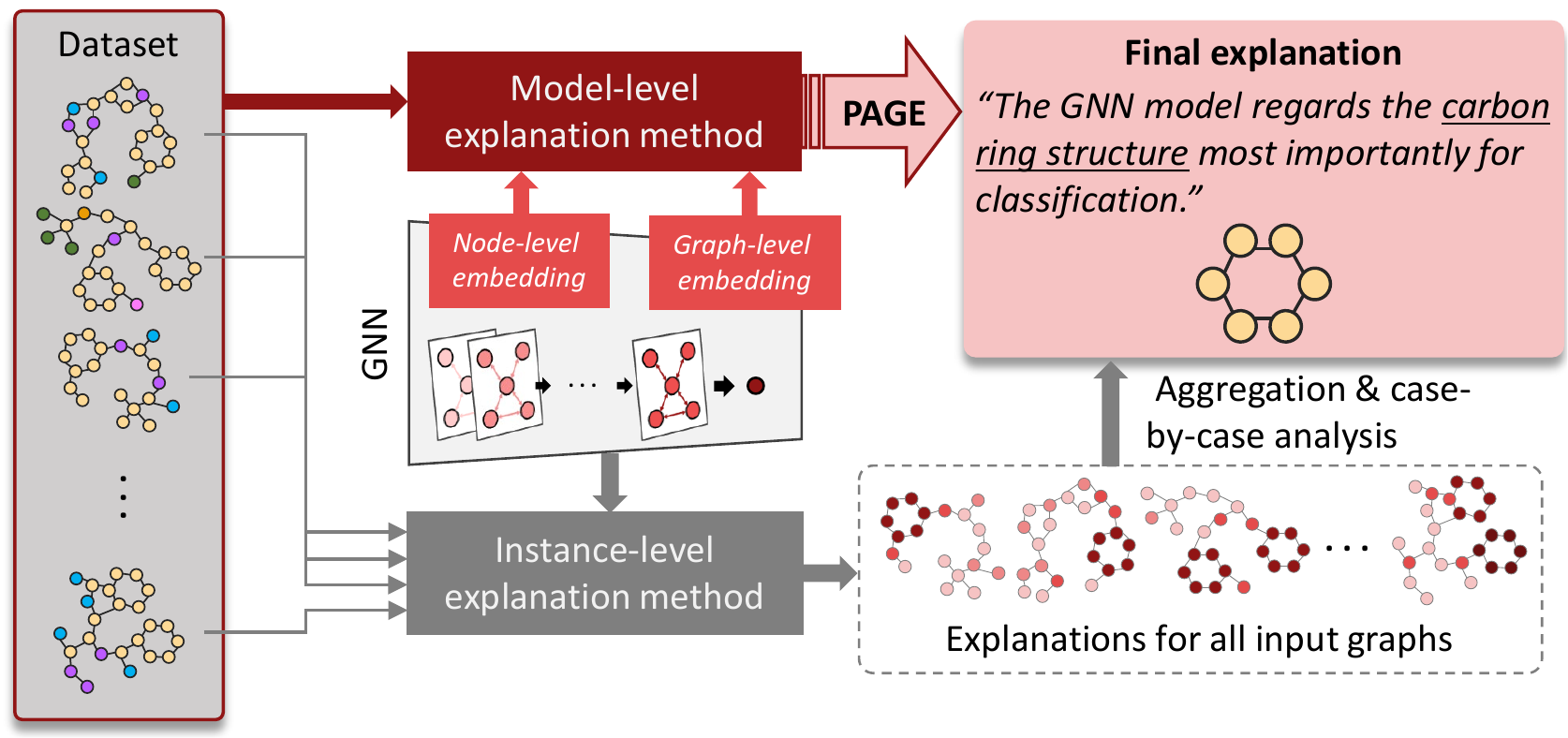}
\caption{A comparison between instance-level and model-level explanation methods to capture the general behavior of GNNs.}
\label{comparisoninstnacevsmodellevel}
\end{figure}

Fig.~\ref{comparisoninstnacevsmodellevel} visualizes how instance-level explanation methods differ from model-level explanation methods in order to \textcolor{black}{understand} the general behavior of the pre-trained GNN model. \textcolor{black}{For instance-level explanations, the explanation is provided for each graph instance. Thus,} to understand the model on the whole dataset, we need further steps such as extraction of common patterns among explanations at the instance-level and aggregation of the extracted patterns~\cite{yuan2020xgnn}. However, model-level explanation methods directly capture how the pre-trained GNN model behaves by presenting what it has learned from the whole dataset\textcolor{black}{, which} is summarized as a form of a human-interpretable subgraph.

\textcolor{black}{As a representative model-level explanation method for GNNs, }XGNN~\cite{yuan2020xgnn} trains a graph generator based on reinforcement learning. Since XGNN requires domain-specific knowledge to manually design the reward function, it often fails to produce reliable explanations unless precise rewards are provided according to different types of datasets. \textcolor{black}{Despite numerous} practical challenges, designing model-level explanation methods for GNNs \textcolor{black}{remains} largely underexplored.

\subsection{Main Contributions}
In this paper, we introduce {\it Prototype-bAsed GNN-Explainer (\textsf{PAGE})}, a novel {\it post-hoc} model-level explanation method that explains what the underlying GNN model has learned for graph classification by discovering human-interpretable {\it prototype graphs}. The GNN model's behavior is represented as the prototype graphs by identifying graph patterns \textcolor{black}{learned by the model} for a given target class. We formulate our problem guided by the following insight: the common graph pattern that we shall seek is indeed within the training set. Thus,  we use the training set as well as representations generated by the pre-trained GNN model \textcolor{black}{as the input to \textsf{PAGE}} in order to precisely capture what the model has learned. Fig.~\ref{comparisoninstnacevsmodellevel} illustrates an example where \textsf{PAGE} correctly identifies the benzene structure from the set of molecules as the resulting model-level explanation, while using a set of both node-level and graph-level embeddings through the pre-trained GNN model as input.

We aim to devise our \textsf{PAGE} method according to the following design principles: an explanation module itself should (a) be interpretable and (b) not be accompanied with advanced learning models (i.e., black-box learning models). Towards this end, we design a two-phase method consisting of 1) clustering and selection of embeddings on the graph-level embedding space and 2) discovery of the prototype graph. In Phase 1, we select class-distinctive graphs from the training set by performing clustering on the graph-level embedding space and then choosing $k$ embeddings near the centroid of each cluster. This phase is motivated by claims that 1) graph-level embeddings manifesting common subgraph patterns tend to be co-located on the graph-level embedding space (which is validated empirically) and 2) the selected $k$ graphs (or equivalently, graph-level embeddings) near the centroid can be good representations of the \textcolor{black}{cluster to which they belong}. In Phase 2, as our explanation, we discover the prototype graph from the selected $k$ graphs. However, representing the GNN's behavior as a precise subgraph pattern from the selected $k$ graphs is challenging since it requires combinatorial search over all possible $k$ nodes across $k$ different graphs\textcolor{black}{, which} should collectively embody what the model has learned. To guide this search, we newly characterize a {\it prototype scoring} function, which is used for efficiently calculating the matching score among $k$ nodes across $k$ selected graphs using node-level embedding vectors. By applying the function to all possible combinations of $k$ nodes, we iteratively search for high matching node tuples until we discover a subgraph. It is worth noting that the impacts and benefits of leveraging the two types of embedding spaces (i.e., node-level and graph-level embedding spaces) in prototype discovery are two-fold: 1) the behavior of the pre-trained GNN model is encoded to vector representations on low-dimensional embedding spaces, which can be handled readily, and 2) the rich attribute and topological information can be fused into vector representations.

To validate the quality and effectiveness of our \textsf{PAGE} method, we \textcolor{black}{perform comprehensive} empirical evaluations using various synthetic and real-world datasets. However, \textcolor{black}{the} evaluation of model-level explanations poses another challenge as it has yet been largely underexplored in the literature. \textcolor{black}{As one of the main contributions of our study}, we device systematic evaluations for model-level GNN explanations, which are summarized below. First, we perform a {\it qualitative} evaluation by visualizing the explanations, which demonstrates that \textsf{PAGE} produces prototype graphs similar to the ground truth explanations. Second, by adopting four performance metrics, accuracy, density, consistency and faithfulness~\cite{sanchez2020gnneval}, with modifications, we carry out the {\it quantitative} analysis, which (a) shows the superiority of \textsf{PAGE} over XGNN~\cite{yuan2020xgnn}, the representative model-level GNN explanation method, and (b) is in agreement with the qualitative assessment. Third, we investigate the relationship between our \textsf{PAGE} method and instance-level explanation methods. To this end, we present two quantities, including the concentration score and relative training gain, and quantitatively assess \textcolor{black}{the extent to which} instance-level explanations agree with the prototype graph discovered by \textsf{PAGE}. Fourth, we analyze the robustness of \textsf{PAGE} by visualizing the resultant prototype graphs in a more difficult setting where each training set is composed of only a few available graphs as \textcolor{black}{the} input of \textsf{PAGE}; we demonstrate that reasonable ground truth explanations can still be produced. Finally, we empirically validate the effectiveness of our prototype scoring function, which successfully approximates computationally expensive alternatives.

Additionally, we address the key differences \textcolor{black}{between} \textsf{PAGE} and XGNN~\cite{yuan2020xgnn}: 1) \textcolor{black}{the use} of {\it interpretable} components without black-box learning modules and 2) {\it generalization} ability in the sense of applying our method to diverse domains without domain-specific prior knowledge. The main technical contributions of this paper are \textcolor{black}{three}-fold and \textcolor{black}{are} summarized as follows:

\color{black}
\begin{itemize}
    \item We propose \textsf{PAGE}, a novel model-level GNN explanation method for graph classification, which is comprised of two phases, 1) clustering and selection of graph-level embeddings and 2) prototype discovery;
    \item In \textsf{PAGE}, we theoretically and empirically justify the usage of clustering of graph-level embeddings and design a new prototype graph discovery algorithm to capture the general behavior of GNN models;
    \item Through our systematic evaluations using five graph classification datasets, we comprehensively demonstrate that \textsf{PAGE} is effective, robust to incomplete datasets, and computationally efficient.
\end{itemize}
\color{black}


\subsection{Organization and Notations}
The remainder of this paper is organized as follows. In Section~\ref{section:relatedwork}, we present prior studies related to our work. In Section~\ref{section:methodology}, we explain the methodology of our study, including the basic settings, description of GNN models, our problem formulation, and an overview of our \textsf{PAGE} method. Section~\ref{section:proposedmethod} describes technical details of the proposed method. Comprehensive experimental results are shown in Section~\ref{section:experiment}. Finally, we provide a summary and concluding remarks in Section~\ref{section:conclusion}.

\section{Related work}\label{section:relatedwork}
We first summarize broader research lines related to explanation methods for deep neural networks, and then we focus on reviewing explanation methods for GNNs.

\subsection{Explanation Methods for Deep Learning}
Significant research efforts have been \textcolor{black}{devoted to} interpretation techniques that explain deep neural network models on image data. \textcolor{black}{Although} existing techniques are commonly partitioned into instance-level and model-level methods, considerable attention has been paid to instance-level explanations \textcolor{black}{, which} explain the prediction \textcolor{black}{of} a given input instance by discovering salient features in the input through the underlying explainable method. As one of widely used techniques, layer-wise relevance propagation (LRP)~\cite{bach2015LRP} \textcolor{black}{proposed to redistribute the model output values towards the input in proportion to the activation values.} \textcolor{black}{In addition}, Grad-CAM~\cite{selvaraju2017gradcam} was presented by producing explanations based on a course heatmap highlighting salient regions in the given image.

In contrast to the instance-level explanations, model-level explanation methods aim at offering the   interpretability of the underlying model itself. These methods generally generate \textcolor{black}{the} optimized input \textcolor{black}{to maximize} a certain prediction score (e.g., a class score) \textcolor{black}{produced by a given explanation model}. DGN-AM~\cite{nguyen2016synthesis} \textcolor{black}{employed a separate neural network capable of inverting the feature representations of an arbitrary layer.} \textcolor{black}{Furthermore,} PPGN~\cite{nguyen2017plugandplay} was developed by adding a learned prior to generate realistic high-resolution images exhibiting more diversity.

\subsection{Explanation Methods for GNNs}

\textcolor{black}{Despite} the great success of GNN models in solving various graph mining tasks\textcolor{black}{,} such as node/graph classification~\cite{wu2019gnnsurvey}, much less attention has been yet paid to the study on {\em explaining} GNN models. For GNN models, explanations \textcolor{black}{in the} form of {\em subgraphs} rather than heatmaps would be appropriate. As a \textcolor{black}{representative work}, GNNExplainer~\cite{ying2019gnnexplainer} was developed to discover a subgraph, including the target node, to be explained by maximizing the mutual information between prediction values of the original graph and the discovered subgraph. GNN-LRP~\cite{schnake2021gnnlrp} extended the idea of LRP to GNN models to produce higher-order explanations via relevant walks contributing to the model decisions. In~\cite{yuan2021subgraphx}, SubgraphX identified the most relevant subgraph explaining the model prediction via Monte Carlo tree search using Shapley values as a measure of subgraph importance. On the other hand, instead of directly searching for subgraphs, several studies \textcolor{black}{have learned} parameterized models \textcolor{black}{to explain} GNN's predictions. PGExplainer~\cite{luo2020pgexplainer} proposed a probabilistic graph generative model to \textcolor{black}{collectively explain multiple instances}. GraphMask~\cite{schlichtkrull2021graphmask} also proposed a post-hoc method that interprets the predictions determining whether (superfluous) edges at every GNN layer can be removed. Furthermore, PGM-Explainer~\cite{vu2020pgmexplainer} presented a probabilistic graphical model so as to provide an explanation by investigating predictions of GNNs when the GNN's input is perturbed. In RC-Explainer~\cite{wang2022rcexplainer}, a reinforcement learning agent was presented to construct an explanatory subgraph by adding a salient edge to connect the previously selected subgraph at each step, where a reward is obtained according to the causal effect for each edge addition. Most recently, CF-GNNExplainer~\cite{lucic2022cfgnnexplainer} presented a counterfactual explanation in the form of minimal perturbation to the input graph such that the model prediction changes.

\textcolor{black}{Although recent efforts have been made to effectively produce {\em instance-level} explanations for GNN models}, {\em model-level} explanations for GNNs have been largely underexplored in the literature. XGNN~\cite{yuan2020xgnn} was developed only for model-level GNN explanations, where a subgraph pattern that maximizes the target class probability is generated via a reinforcement learning framework. It is worth mentioning that several explanation methods that provide more coarse-grained explanations were \textcolor{black}{recently} proposed. ReFine~\cite{Wang2021Multigrained} proposed to train global attribution that captures class-wise patterns, while also training local attribution by fine-tuning on the given instance. CGE~\cite{Fang2023coopxai} jointly considered attribution on the input as well as the intermediate neural activation in the underlying GNN model for explanations, and found a large overlap of neural activations within instances from the same class. However, the class-wise patterns still require a specific instance as input to be expressed, and cannot map such explanations as a standalone structural representation, which is required in model-level explanation methods.
\color{black}

\subsection{Discussion}

Apart from most explanation methods \textcolor{black}{that operate on the instance-level, } we aim \textcolor{black}{to design} model-level interpretations that \textcolor{black}{offers high-level insights as well as generic understandings of the underlying model mechanism}. Critically, XGNN requires domain knowledge to provide appropriate rewards in the graph generation procedure based on reinforcement learning. If the reward is not adequately designed, then the generated representative graph would be hardly realistic. Furthermore, the usage of such reinforcement learning frameworks \textcolor{black}{introduces} another black-box model for explanation, which is a sub-optimal choice. This motivates us to design a more interpretable and convenient model-level method \textcolor{black}{that does not require} domain-specific knowledge and learning modules.

We note that the prototype discovery in Phase 2 of our \textsf{PAGE} method can be viewed as a form of graph matching~\cite{Yan2016graphmatching, Fey2020graphconsensus, Sole-Ribalta2013graphmatching, Wang2020graphmatching}, since graph matching attempts to compare common substructures among multiple graph instances. Although graph matching methods can be utilized in Phase 2 to find the prototype candidates, they require a substantial amount of nontrivial modifications since they cannot incorporate the instance-level embedding vectors, nor can they guarantee that a graph with a single connected component is returned, which thereby severely reduces the quality of the discovered prototype candidates.

\color{black}


\section{METHODOLOGY} \label{section:methodology}
In this section, we first describe our problem setting with the objective. Then, we present the overview of our \textsf{PAGE} method as a model-level explanation for GNNs.

\begin{figure*}[t]
\centering
\begin{subfigure}{\textwidth}
  \centering  \includegraphics[width=0.85\textwidth]{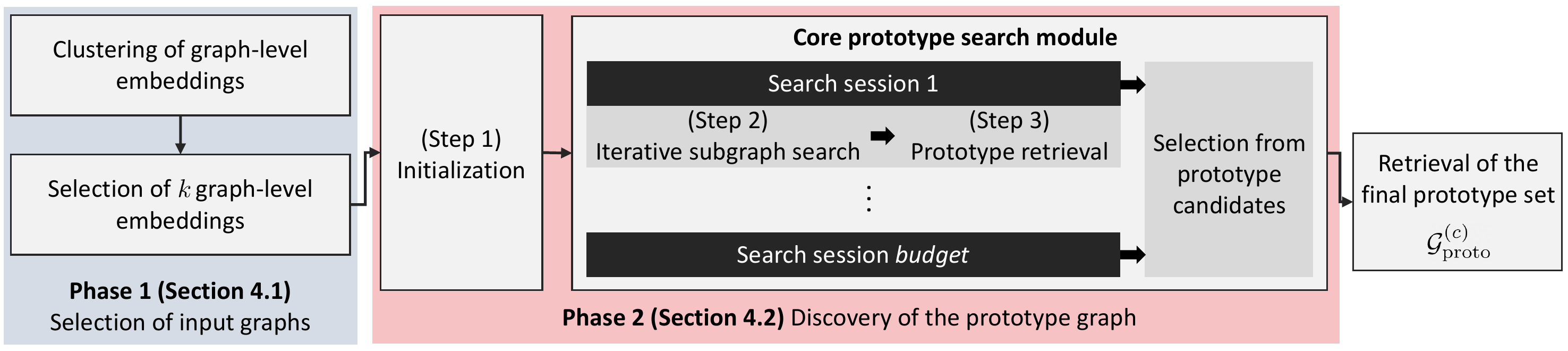}   
  \caption{A schematic overview of our \textsf{PAGE} method.} \label{endtoendPAGE}
\end{subfigure}

\begin{subfigure}{0.38\textwidth}
  \centering
  \includegraphics[width=\linewidth]{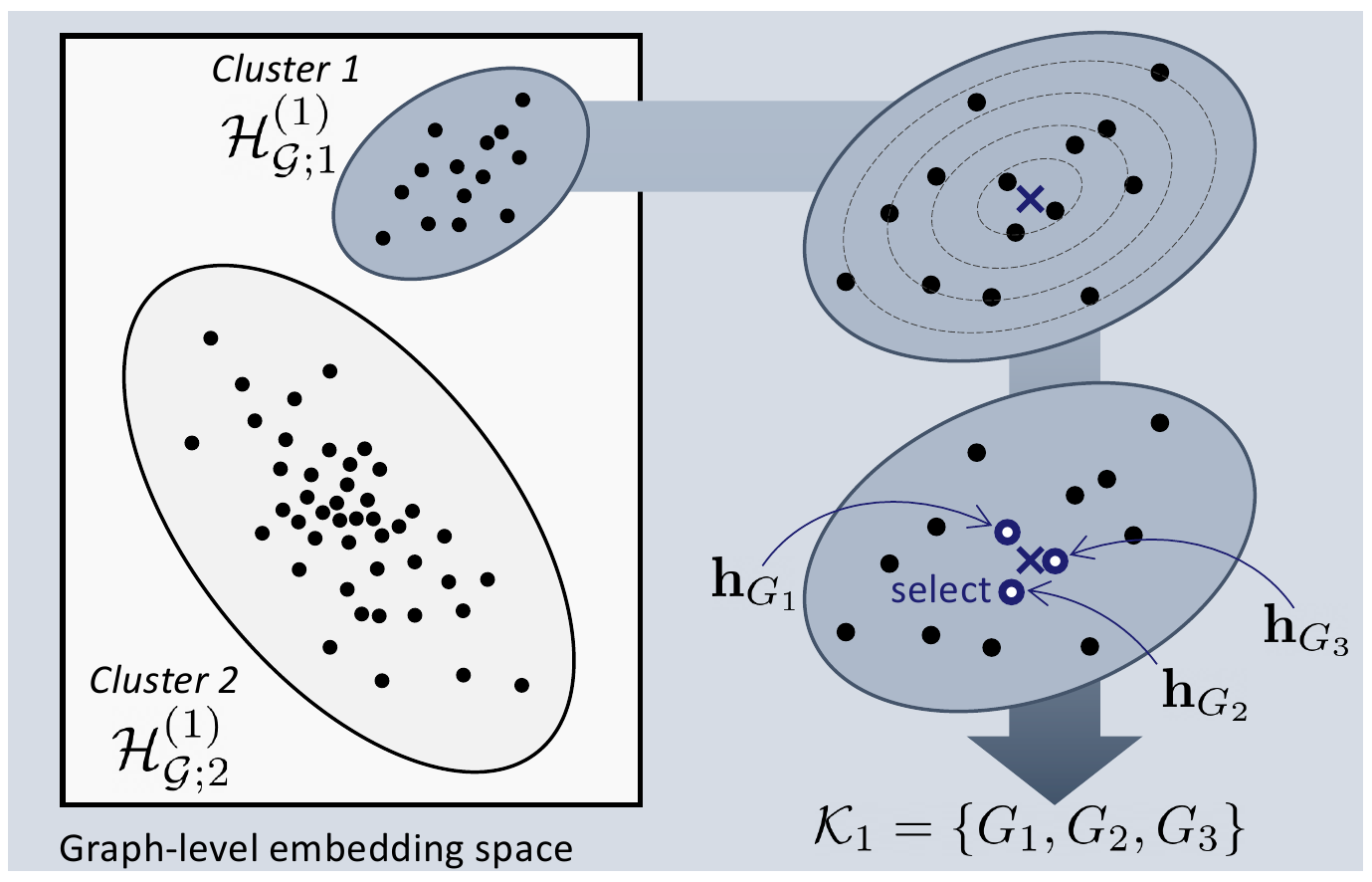}  
  \caption{Phase 1 of \textsf{PAGE} when $k=3$.}
  \label{figure:phase1}
\end{subfigure}
\begin{subfigure}{0.55\textwidth}
  \centering
  \includegraphics[width=\linewidth]{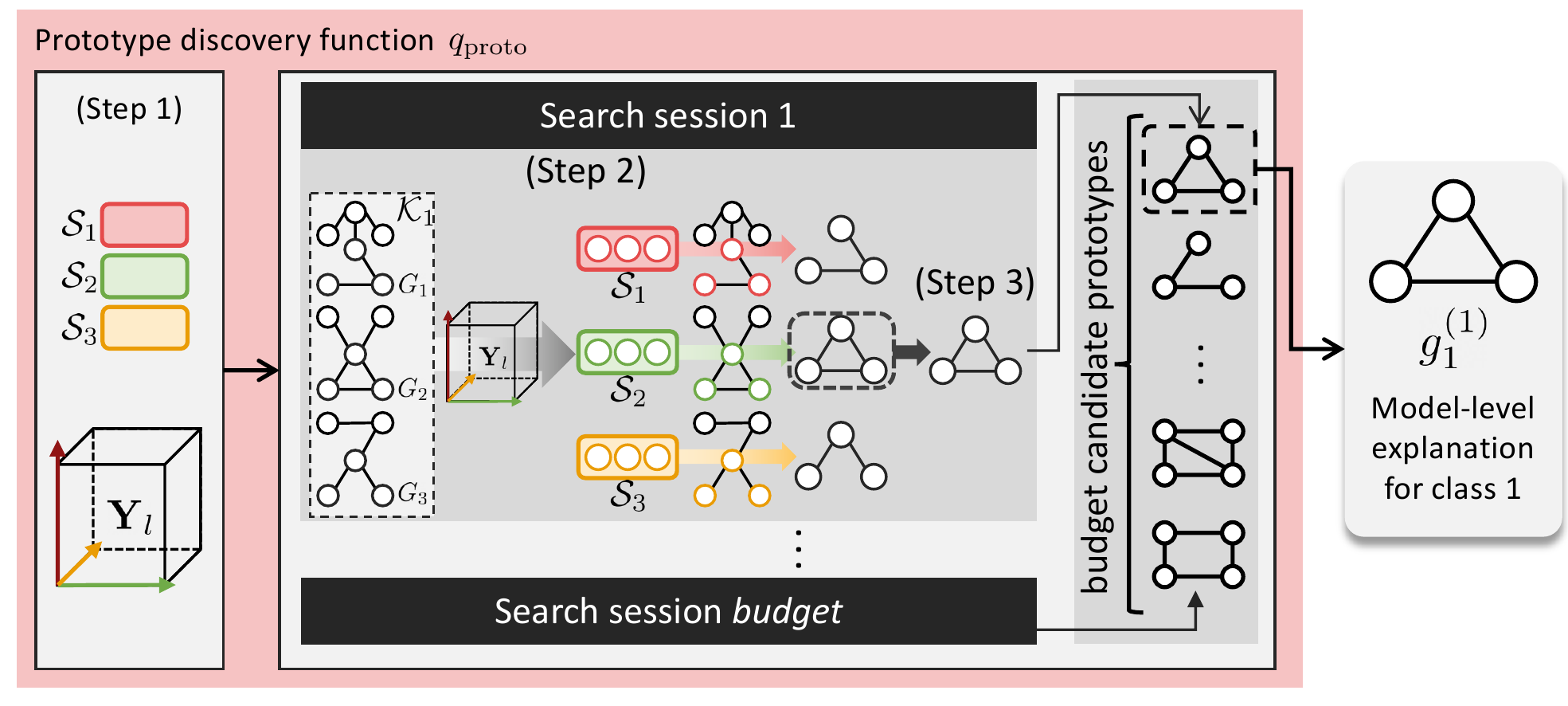}  
  \caption{Phase 2 of \textsf{PAGE} when $k=3$.}
  \label{figure:phase2}
\end{subfigure}
\caption{A schematic overview overview and an illustrative example of our \textsf{PAGE} method.}
\end{figure*}

\subsection{Settings and Basic Assumptions}
Let us assume that we are given a set of $n$ graphs $\mathcal{G} = \{G_i\}_{i=1}^{n}$ with node attributes. Each graph is denoted as $G_i = (\mathcal{V}_i, \mathcal{E}_i, \mathcal{X}_i)$, where $\mathcal{V}_i = \{v_i^1, \cdots, v_i^{|\mathcal{V}_i|}\}$ is the set of nodes with cardinality $|\mathcal{V}_i|$ and $\mathcal{E}_i$ is the set of edges between pairs of nodes in $\mathcal{V}_i$. We assume each graph $G_i$ to be undirected and unweighted without self-edges or repeated edges. We define $\mathcal{X}_i = \{{\bf x}_{i}^{1}, \cdots, {\bf x}_{i}^{|\mathcal{V}_i|}\}$ as the set of feature (attribute) vectors of nodes in $\mathcal{V}_i$, where ${\bf x}_i^j \in \mathbb{R}^{d}$ is the feature vector of node $v_i^j$ in $G_i$ and $d$ is the number of features per node. 

We associate each graph $G_i$ with a label $y_i$, where $y_i$ is an element in the set $\mathcal{C} = \{c_1, \cdots, c_m\}$, which means that we \textcolor{black}{address} an $m$-class graph classification problem. \textcolor{black}{A} pre-trained GNN model $f_{\text{GNN}}: \mathcal{G} \rightarrow \mathcal{C}$ for graph classification corresponds to the model of interest to be explained. Additionally, the number of ground truth explanations per class is assumed to be available for each dataset. While acquiring such knowledge a priori from each dataset may \textcolor{black}{require} additional work, we do not include this task in our study to focus primarily on prototype discovery.

\subsection{GNN Models for Graph Classification}\label{section:preliminary}
In this subsection, we give a brief review of GNN models for graph classification. 

First, we show a general form of message passing in GNNs\textcolor{black}{~\cite{gilmer2017mpnn}}, in which we iteratively update the representation of a node by aggregating representations of its neighbors using two functions. Specifically, given a graph $G_i$, the underlying GNN model produces the latent representation vector ${\bf u}_{v}^{p}$ of node $v \in \mathcal{V}_i$ at the $p$-th GNN layer, where $p \in \{1, \cdots, P\}$.\footnote{To simplify notations, $v_{i}^{j}$ will be written as $v$ unless dropping the superscript $j$ and the subscript $i$ causes any confusion.} Formally, the $p$-th layer of the GNN model updates the latent representation ${\bf u}_{v}^{p}$ of each node $v$ in $\mathcal{V}_i$ by passing through two phases, including the message passing phase and the update phase, which are expressed as ${\bf m}_{v}^{p+1} = \sum_{v' \in \mathcal{N}_{G_i}(v)} M_{p}({\bf u}_{v}^{p} {\bf u}_{v'}^{p})$ \textcolor{black}{and} ${\bf u}_{v}^{p+1} = U_{p}({\bf u}_{v}^{p}, {\bf m}_{v}^{p+1}),$
respectively, where $\mathcal{N}_{G_i}(v)$ indicates the set of neighbors of node $v$ in the given graph $G_i$. Here, the message function $M_{p}(\cdot, \cdot)$ and the update function $U_{p}(\cdot, \cdot)$ can be specified by several types of GNN models.\footnote{As an example, GCN~\cite{kipf2017gcn} can be implemented by using $M_{p}({\bf u}_{v}^{p}, {\bf u}_{w}^{p}) = (\sqrt{\text{deg}(v)\text{deg}(w)})^{-1/2}{\bf u}_{v}^{p}$ and $U_{p} = \sigma(W^{p} \cdot {\bf m}_{v}^{p+1})$, where $\text{deg}(\cdot)$ is the degree of each node and $\sigma(\cdot)$ is the activation function.} After passing through all $P$ layers, we acquire the final representation as the {\em node-level} embedding vector of node $v_i^j$, which is denoted as ${\bf h}_{i}^{j} = {\bf u}_{v_{i}^{j}}^{P}\in\mathbb{R}^{b}$ for the dimension $b$ of each vector.

\textcolor{black}{Second, we describe how to produce the \textit{graph-level} embedding vector ${\bf h}_{G_i}$ of $G_i$. To this end, a readout function $R(\cdot)$ collects all node-level embedding vectors ${\bf h}_{i}^{j}$ for $v_{i}^{j} \in \mathcal{V}_i$ and is formally expressed as }${\bf h}_{G_{i}} = R(\mathcal{H}_{G_i}),$
where $\mathcal{H}_{G_i} = \{{\bf h}_{i}^{j}|v_{i}^{j} \in \mathcal{V}_{i}\}$, indicating the set of embedding vectors of nodes in $\mathcal{V}_{i}$. \textcolor{black}{In practice}, differentiable permutation-invariant functions such as the summation and average~\cite{xu2019gin} are typically used. Then, the set of calculated graph-level embedding vectors over all graphs $\mathcal{G}$ is fed into a classifier to perform graph classification.

\subsection{Problem Formulation}
The objective of our study is to provide a post-hoc {\em model-level} explanation by discovering the most distinctive graph patterns that the underlying GNN model has learned during training for graph classification. Towards this end, we present \textsf{PAGE}, a new model-level explanation method along with human-interpretable {\em prototype} graphs. For a target class $c \in \mathcal{C}$ to which a given graph belongs and a pre-trained GNN model $f_{\text{GNN}}$, \textsf{PAGE} aims to discover a set of prototype graphs, $\mathcal{G}_{\text{proto}}^{(c)}$, each of which represents the graph pattern such that the model $f_{\text{GNN}}$ most likely predicts the target class $c$. Formally, the model-level explanation function $f_{\text{model}}^{X}(\cdot, \cdot)$ in \textsf{PAGE} is expressed as:
\begin{equation}
    \mathcal{G}_{\text{proto}}^{(c)} = f_{\text{model}}^{X}(c ; f_{\text{GNN}}).
\end{equation}

Note that the model-level explanation differs from the instance-level explanation, which provides an explanation with respect to a given graph instance $G_i \in \mathcal{G}$ rather than a target class. 

\subsection{Overview of Our \textsf{PAGE} Method}
In this subsection, we explain our methodology along with the overview of the proposed \textsf{PAGE} method, which \textcolor{black}{consists} of the following two phases: 1) selection of input graphs that precisely represent the given class $c$ through clustering and selection of graph-level embeddings, and 2) discovery of the prototype graph set $\mathcal{G}_{\text{proto}}^{(c)}$.

Before describing each phase above, let us briefly state the core difference from XGNN~\cite{yuan2020xgnn}, the state-of-the-art model-level explanation method for GNNs. While XGNN was designed to {\em generate} the prototype graph, \textsf{PAGE} attempts to {\em discover} the set of prototype graphs, $\mathcal{G}_{\text{proto}}^{(c)}$, using underlying graphs, $\mathcal{G}$, by turning our attention to both the graph- and node-level embedding spaces, where the impacts and benefits of leveraging the two types of embedding spaces are two-fold. First, the behavior of the pre-trained GNN model is encoded to vector representations on low-dimensional embedding spaces, thus facilitating many network analysis tasks. Second, the rich attribute and structural information of $\mathcal{G}$ can be \textcolor{black}{fused efficiently and effectively} into vector representations in conducting the prototype discovery. In particular, the set of graph-level embedding vectors can be used to reduce the prototype graph search space within the set $\mathcal{G}$ by selecting only a small number of embedding vectors that are most representative.

Now, we describe each phase of our \textsf{PAGE} method as follows. First, we select a small number of graphs from the input that represent the given class $c$ by performing clustering on the graph-level embedding space (see the left part in Fig.~\ref{endtoendPAGE}). This phase is motivated by the observation that graph-level embedding vectors revealing common subgraph patterns tend to be co-located on the graph-level embedding space (which will be empirically shown in Section~\ref{subsubclustertheoryempirical}). The clustering phase also provides a practical advantage, where it provides a natural anchor point to which we select graph-level embedding vectors (and their corresponding graphs) as the centroid of each cluster. For a target class $c$, we focus on $\mathcal{H}_{\mathcal{G}}^{(c)}$, which is defined as a subset of graph-level embedding vectors such that the model $f_{\text{GNN}}$ predicts as class $c$. In this phase, we discover $L$ clusters using a Gaussian mixture model (GMM)~\cite{bishop2006prml} for a pre-defined parameter $L$. \textcolor{black}{Consequently, we are able to map} each graph-level embedding vector to one of $L$ clusters. After acquiring $L$ clusters, we select the $k$-nearest neighbors ($k$NNs) from each cluster's centroid on the graph-level embedding space for a pre-defined parameter $k$. Since centroids are typically \textcolor{black}{located near the} center of each cluster, graph-level embedding vectors near centroids can be \textcolor{black}{considered} good representations of \textcolor{black}{the clusters to which they belong}. \textcolor{black}{Subsequently}, we acquire $\mathcal{K}_{l}$, which represents the subset of underlying graphs in $\mathcal{G}$ corresponding to $k$ selected graph-level embedding vectors. That is, we retrieve a subset of graphs from the selected graph-level embedding vectors. Fig.~\ref{figure:phase1} illustrates an example of retrieving $\mathcal{K}_1 = \{G_1, G_2, G_3\}$ using the 3-nearest neighbors ${\bf h}_{G_1}, {\bf h}_{G_2}$, and ${\bf h}_{G_3}$ from the centroid when $k=3$.

Second, we turn to discovering the prototype graph (see the right part in Fig.~\ref{endtoendPAGE}). For the ease of explanation, we focus only on the $l$-th cluster. Given $\mathcal{K}_l = \{G_1, \cdots, G_k\}$ for the $l$-th cluster, we aim to discover a common subgraph pattern within the subset $\mathcal{K}_l$, which serves as the prototype graph, denoted as $g_l^{(c)}$, in the set $\mathcal{G}_{\text{proto}}^{(c)}$. To this end, we compare nodes across $k$ different graphs in $\mathcal{K}_l$ in order to select the most probable nodes according to the {\em matching score}. The matching score, which is calculated among $k$ nodes, takes advantage of the structural and attribute information embedded by the model $f_{\text{GNN}}$, which measures how high the matching score for the node-level embedding vectors of nodes $v_1^{i_1},\cdots,v_k^{i_k}$ is. After we use the matching scores to select the set of nodes to be included in the explanation, we extract $k$ induced subgraphs from $\mathcal{K}_l$ by only taking into account the selected nodes. \textcolor{black}{We can then calculate} the output probability $p_\text{GNN}$ for each subgraph extracted from $k$ graphs in the set $\mathcal{K}_l$ by feeding each into $f_{\text{GNN}}$, where the subgraph with the highest $p_\text{GNN}$ is selected. We further repeat this procedure with a different search process during multiple search sessions alongside a given {\it search budget}. Finally, the subgraph exhibiting the highest $p_\text{GNN}$ among the ones found across all search sessions is retrieved as the resulting prototype graph, which provides an intuitive model-level explanation for GNN's prediction given the target class $c$. Fig.~\ref{figure:phase2} illustrates an example of retrieving the final prototype graph $g_1^{(1)}$ based on the prototype discovery function $q_\text{proto}$ using $\mathcal{K}_1=\{G_1,G_2,G_3\}$.

\section{\textsf{PAGE}: Proposed Method} \label{section:proposedmethod}
In this section, we elaborate on \textsf{PAGE}, the proposed post-hoc model-level GNN explanation method for graph classification. We first describe how to cluster and select graph-level embeddings for the subset selection of input graphs in detail. We also provide the theoretical and empirical validation for our clustering. Then, we present how to discover prototype graphs. 

\subsection{Clustering and Selection of Graph-Level Embeddings (Phase 1)} \label{mainsection:clusteringandselection}
\subsubsection{Methodological Details} \label{mainsection:clusteringandselectionsub}

We assume that \textcolor{black}{the} parameters of a GNN model $f_\text{GNN}$ are trained by a set of graphs, $\mathcal{G}$. Given a target class $c \in \mathcal{C}$, we obtain a set of graph-level embedding vectors, $\mathcal{H}_\mathcal{G}^{(c)}$, through a feed-forward process of GNN.

Then, we estimate the parameters of the Gaussian mixture distribution to fit the embedding vectors $\mathcal{H}_\mathcal{G}^{(c)}$. More precisely, we apply an expectation maximization (EM) algorithm~\cite{dempster1977em} to estimate the parameters $ \{ ( \pi_l, {\bm \mu}_l, {\mathbf \Sigma}_l ) \}_{l=1}^{L}$ with the mean vector ${\bm \mu}_l$ and the covariance matrix ${\mathbf \Sigma}_l$ for the $l$-th cluster of the Gaussian mixture distribution $p({\bf h}) = \sum_{l=1}^{L}\pi_l\mathcal{N}({\bf h} | {\bm \mu}_l, {\bm \Sigma}_l)$ \textcolor{black}{due to the theoretical guarantees of monotonically increasing likelihood and convergence~\cite{Daskalakis2017EM} as well as the robustness to noisy input samples~\cite{SAMMAKNEJAD2019123}.\footnote{\textcolor{black}{Although the convergence of EM is slow (i.e., linear convergence) in approaching the true parameters, we empirically showed that the runtime of the EM algorithm in Phase 1 is not a bottleneck.}}} The EM iteration alternates between performing the E-step and M-step as follows. In the E-step, we calculate $\gamma^{(t)}_{nl}$, which represents the posterior probability of the $n$-th vector in $\mathcal{H}_{\mathcal{G}}^{(c)}$ being assigned to the $l$-th cluster and is expressed as \color{black}$\gamma^{(t)}_{nl} = \dfrac{\pi_{l}\mathcal{N}({\bf h}_{G_n}|{\bm \mu}^{(t)}_{l}, {\bf \Sigma}^{(t)}_{l})}{\sum_{i=1}^{L}\pi_{i}\mathcal{N}({\bf h}_{G_n} | {\bm \mu}^{(t)}_{i}, {\bf \Sigma}^{(t)}_{i})}$,\color{black} 
where the superscript $(t)$ indicates the EM iteration index. In the M-step, we re-estimate the GMM parameters, \color{black}${\bm \mu}_{l}^{(t+1)} = \dfrac{1}{N_{l}}\sum_{n=1}^{|\mathcal{H}_{\mathcal{G}}^{(c)}|}\gamma^{(t)}_{nl}{\bf h}_{G_n}$, ${\bf \Sigma}_{l}^{(t+1)} = \dfrac{1}{N_{l}}\sum_{n=1}^{|\mathcal{H}_{\mathcal{G}}^{(c)}|}\gamma^{(t)}_{nl}({\bf h}_{G_n} - {\bm \mu}_{l}^{(t+1)})({\bf h}_{G_n} - {\bm \mu}_{l}^{(t+1)})^{T}$, \color{black}
where $N_{l} = \sum_{n=1}^{|\mathcal{H}_{\mathcal{G}}^{(c)}|}\gamma(z_{nl})$. Using the estimated parameters, we assign each graph-level embedding vector ${\bf h}_{G_i}$ to the cluster with the highest probability $\gamma^{(t)}_{nl}$. In other words, we divide $\mathcal{H}_{\mathcal{G}}^{(c)}$ into $L$ disjoint subsets $\mathcal{H}_{\mathcal{G};1}^{(c)},\cdots, \mathcal{H}_{\mathcal{G};L}^{(c)},$ where $\mathcal{H}^{(c)}_{\mathcal{G};l}$ contains graph-level embedding vectors assigned to the $l$-th cluster by the GMM. 

Next, we select the $k$NNs from each cluster's centroid on the graph-level embedding space in order to find the set of $k$ graphs, $\mathcal{K}_l$, for each cluster. \textcolor{black}{Since the covariance matrix is available for each cluster, we are able to utilize the Mahalanobis distance that takes into account both the correlations and the shape of each cluster on the graph-level embedding space.} We sort the graph-level embedding vectors ${\bf h}_{G_i}$ in descending order with respect to the Mahalanobis distance from the mean vector ${\bm \mu}_l$ within the $l$-th cluster~\cite{bishop2006prml}: \color{black}$(({\bf h}_{G_i} - {\bm \mu}_{l})^{T}{\bf \Sigma}_{l}^{-1}({\bf h}_{G_i} - {\bm \mu}_{l}))^{1/2}.$\color{black}
Then, for each cluster, we select $k$ graph-level embedding vectors that are closest to ${\bm \mu}_{l}$ and their corresponding input graphs $\mathcal{K}_l$ in $\mathcal{G}$.

\subsubsection{Theoretical and Empirical Validation} \label{subsubclustertheoryempirical}
In Section~\ref{subsubclustertheoryempirical}, we aim at justifying the usage of clustering of graph-level embeddings in \textsf{PAGE} via our theoretical and empirical validation. Our study basically presumes that graphs containing different prototypes are highly likely to be mapped into different vector representations on the graph-level embedding space, even though they belong to the same class. To validate this claim, we first provide a theoretical foundation that connects GNNs and the {\em Weisfeiler-Lehman (WL) kernel}\textcolor{black}{~\cite{morris2019wlgoneural}}. We then empirically validate the above claim by visualizing graph-level embeddings for a benchmark dataset.

We \textcolor{black}{begin} by describing the WL kernel\textcolor{black}{~\cite{morris2019wlgoneural}}. It initially colors each node of a given graph by the node attribute and iteratively refines the node coloring in rounds by taking into account the colors in the neighbors of each node until stabilization. We formally state this by assuming that a graph $G_i$ is given along with the initial color for each node. A node coloring $c^{(p)}$ is a function that bijectively maps each node $v\in\mathcal{V}_i$ in $G_i$ to a unique color that has not been used in previous iterations, where $p \geq 0$ indicates the iteration index. For the target node $v$, the WL kernel iteratively updates $c^{(p)}(v)$ by recoloring over the multiset of neighborhood colors, i.e., $\{\!\{ c^{(p-1)}(v') | v'\in\mathcal{N}_{G_{i}}(v)\}\!\}$, where $\mathcal{N}_{G_i}(v)$ indicates the set of neighbors of node $v$ in $G_i$.\footnote{A multiset is a generalized version of a set that allows multiple instances for each of its elements, where the multiplicity of each element in the multiset corresponds to the number of instances.} Now, we are ready to establish the following theorem, which verifies that there exists a GNN model equivalent to the WL kernel as long as node-level embeddings are concerned.

\begin{theorem}[Morris et al., 2019]~\label{theorem:morris}
Let $f_\text{GNN}^{(p)}$ denote the function that returns node-level vector representations at the $p$-th GNN layer in $f_\text{GNN}$. Then for all $p \geq 0$, there exists a GNN model {\em equivalent} to a node coloring such that $f_\text{GNN}^{(p)}(u)=f_\text{GNN}^{(p)}(v)$ if and only if $c^{(p)}(u)=c^{(p)}(v)$ for every $u,v\in\mathcal{V}_i$.
\end{theorem}

From Theorem~\ref{theorem:morris}, the node-level embedding vector ${\bf h}_i^j$ of node $v_i^j$ is equivalent to $c^{(P)}(v_i^j)$. Next, we \textcolor{black}{explore} the relationship between a GNN model and the WL kernel in discovering {\em graph-level} embeddings. To this end, we first acquire a graph feature vector of $G_i$, denoted as ${\bf l}_{G_i} = [n_i^1, \cdots, n_i^s]$, where $s$ is the number of unique node colors and $n_i^l$ is the number of nodes for which color $l$ has been assigned for $l=1,\cdots,s$. The graph feature vector ${\bf l}_{G_i}$ is used to compute similarities between pairs of graphs. The following corollary states the equivalence between the GNN model and the WL kernel in terms of graph-level embeddings.

\begin{corollary}~\label{colrollary:graphrepresentation}
Suppose that two graph feature vectors ${\bf l}_{G_a}$ and ${\bf l}_{G_b}$ are obtained via color refinement for two graphs $G_a$ and $G_b$, respectively. Then, for a GNN model with $P$ layers such that $c^{(P)}$ and $f_\text{GNN}^{(P)}$ are equivalent, ${\bf l}_{G_a} = {\bf l}_{G_b}$ implies $R(\mathcal{H}_{G_a}) = R(\mathcal{H}_{G_b})$, when $R(\cdot)$ is either the summation or average function.
\end{corollary}

\begin{figure}[t]
\centering
\includegraphics[width=0.85\columnwidth]{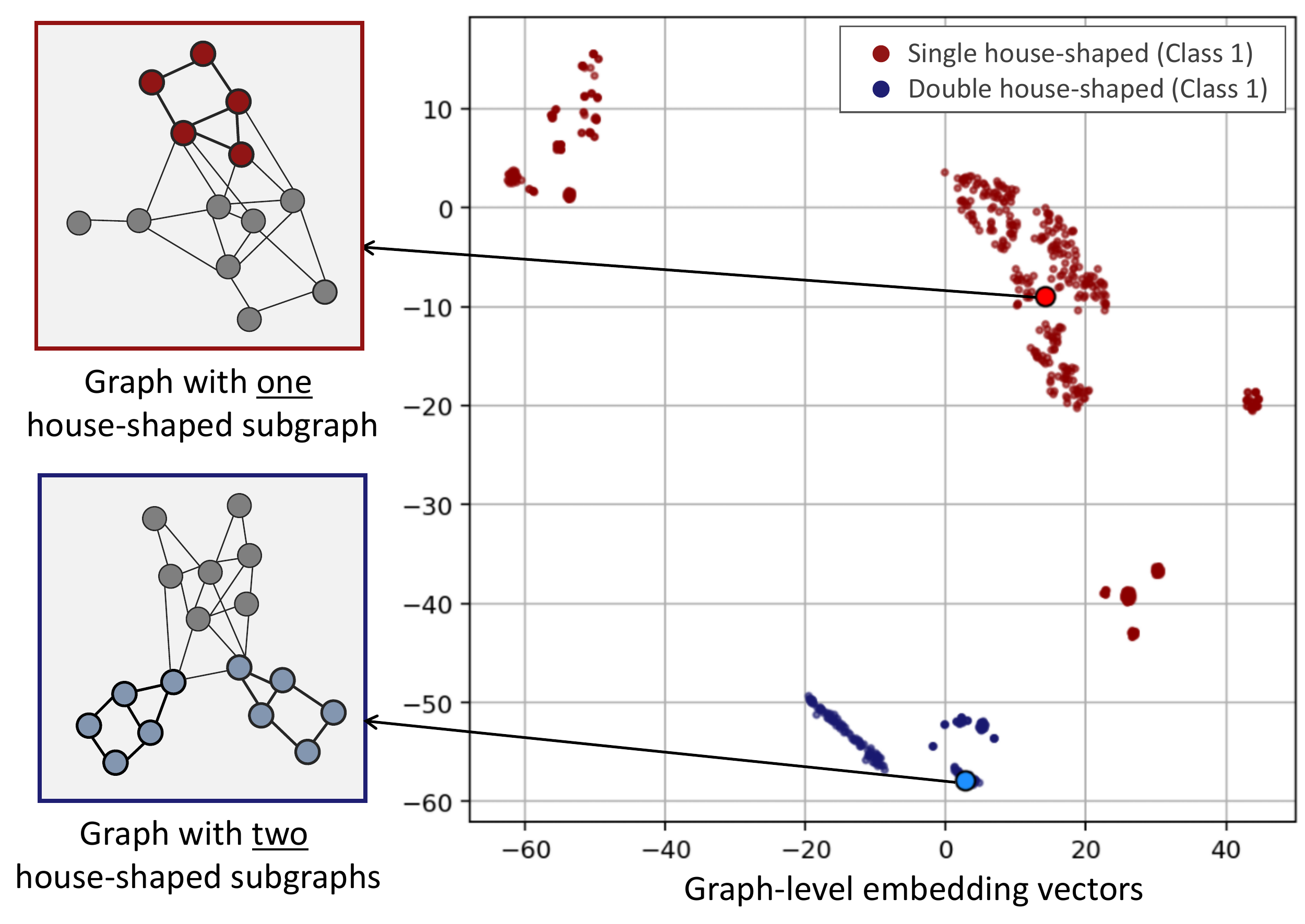}
\caption{Visualization of graph-level embeddings for the BA-house dataset.}
\label{figure:clusteringexample}
\end{figure}

\begin{proof}
From Theorem~\ref{theorem:morris}, there exists a bijective function $\nu: c^{(P)}(v_i^j) \rightarrow {\bf h}_i^j$ for node $v_i^j\in \mathcal{V}_i$, where $i$ is {\em either} $a$ or $b$. Moreover, ${\bf l}_{G_a} = {\bf l}_{G_b}$ indicates that both the number of nodes and the distribution of node colors over each graph are the same for $G_a$ and $G_b$. Due to the fact that each node-level embedding vector ${\bf h}_i^j$ can be regarded as a bijection $\nu$ of a node color, the distributions of unique node-level embedding vectors for $G_a$ and $G_b$ are identical. In this context, if the readout function $R(\cdot)$ is the summation, then we have
\begin{equation} \label{corollaryeq}
\sum_{j=1}^{|\mathcal{V}_a|}{\bf h}_a^j = \sum_{j'=1}^{|\mathcal{V}_b|}{\bf h}_b^{j'},
\end{equation}
thus resulting in $R(\mathcal{H}_{G_a})=R(\mathcal{H}_{G_b})$, since the summation is permutation-invariant. When $R(\cdot)$ is given by the average function, this corollary can also be proven by dividing each side of (\ref{corollaryeq}) by the number of nodes. This completes the proof of this corollary.
\end{proof}

From Corollary~\ref{colrollary:graphrepresentation}, one can expect that GNNs will behave as WL kernels in the sense of differentiating graphs (i.e., distinguishing non-isomorphic graphs).

\begin{figure*}[t]
  \centering  \includegraphics[width=0.85\textwidth]{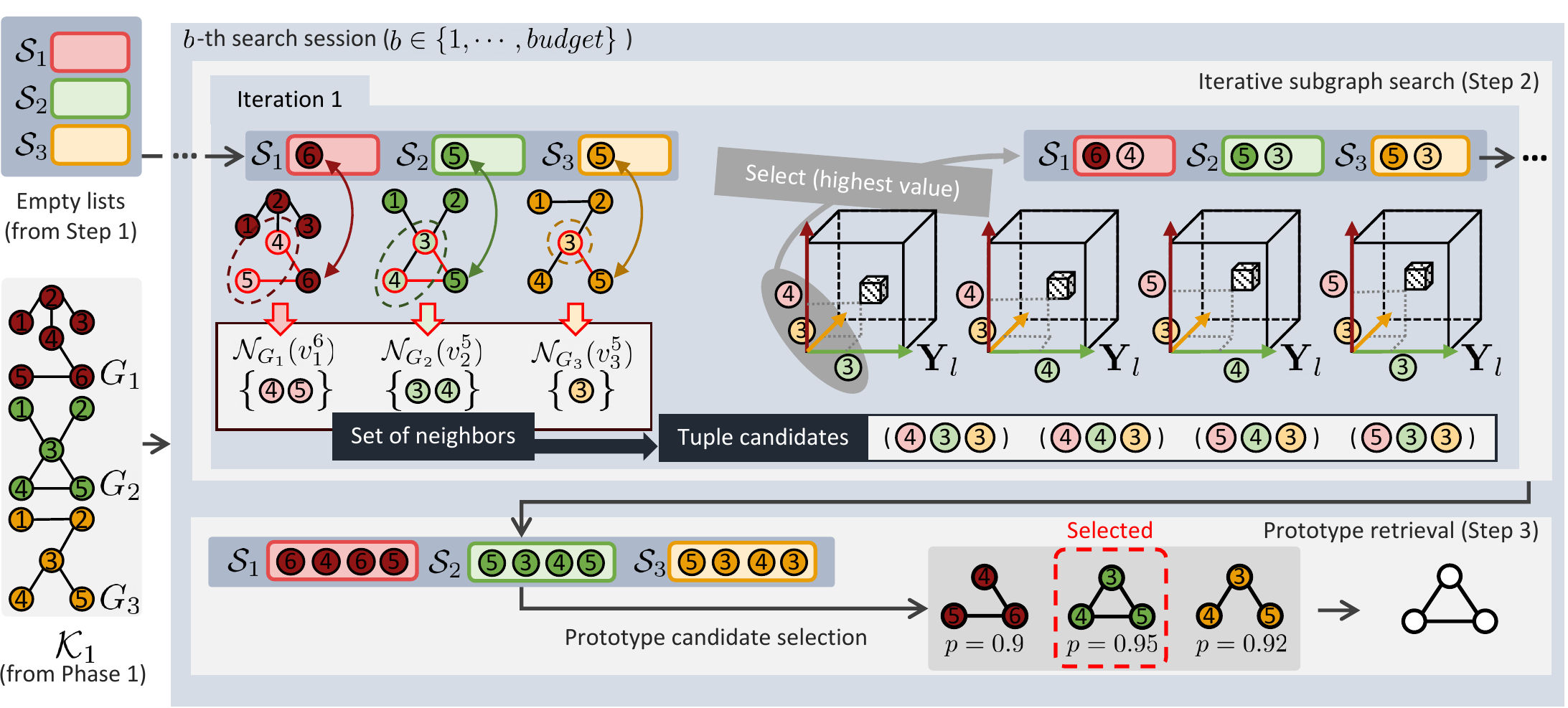}
  \caption{An example that illustrates the mechanism of the core prototype search module for a certain search session when $k=3$.}
\label{figure:mainiteration}
\end{figure*}

Next, we empirically validate our claim that GNNs can differentiate graphs containing different prototypes.
\begin{example}
\textcolor{black}{Fig.~\ref{figure:clusteringexample} empirically demonstrates such a tendency by visualizing the graph-level embedding vectors for the BA-house dataset via $t$-SNE~\cite{vandermaaten2008tsne} in which there are multiple graphs with and without the house-shaped subgraph(s)} (refer to Section~\ref{subsection:datasets} for more description of BA-house). Here, we train a GNN model to perform a graph classification task that classifies whether an input graph contains the house-shaped subgraph(s) (Class 1) or not (Class 0). However, \textcolor{black}{as} each graph in Class 1 may have either one or two house-shaped subgraphs, we are interested in further investigating whether graphs with different numbers of house-shaped subgraphs, belonging to Class 1, are separately embedded into the graph-level embedding space. Although the GNN model is trained while being unaware of the existence of such prototypes, we observe that the GNN model embeds input graphs with different prototypes to distant clusters in the graph-level embedding space (see red and blue points in Fig.~\ref{figure:clusteringexample}).
\end{example}
Through this empirical validation, we conclude that GNNs are indeed capable of differentiating graphs with different prototypes as in the WL kernel.

\subsection{Prototype Discovery (Phase 2)} \label{mainsection:prototypediscovery}
In this section, we turn our attention to describing our prototype discovery algorithm $q_{\text{proto}}$, which calculates the matching score by comparing $k$ nodes across $k$ different graphs in the set $\mathcal{K}_l = \{G_1, \cdots, G_k \}$. Since we iteratively discover a prototype graph $g_{l}^{(c)}$ for each cluster $l\in\{1,\cdots,L\}$, we concentrate only on obtaining $g_{l}^{(c)}$ for the $l$-th cluster in this subsection. The prototype discovery phase consists of the initialization step and the core prototype search module that is made up of two steps. The mechanism of the core prototype search module for a certain search session is illustrated in Fig.~\ref{figure:mainiteration} where the parameter $k$ in $k$NN is given by 3.

\begin{remark}
It is possible to employ alternatives (e.g., using graph matching) to achieve our goal of discovering prototype graphs. Nonetheless, when we adopted the graph matching method in~\cite{Sole-Ribalta2013graphmatching, Wang2020graphmatching} to discover prototype graph candidates, we empirically found that the resulting candidates are less likely to include the ground truth explanation, if not unrecognizable at all. This implies that prototype discovery based on alternative approaches may require major modifications \textcolor{black}{for satisfactory performance}.
\end{remark}

\subsubsection{Initialization (Step 1)} \label{section:matchingtensor}
The initialization step \textcolor{black}{involves} the creation of an order-$k$ matching tensor ${\bf Y}_l$ and the creation of empty node lists. We first describe how to create ${\bf Y}_l$. From the set of node-level embedding vectors, $\mathcal{H}_{G_i} \in \mathbb{R}^{|\mathcal{V}_i| \times b}$ for each $G_i \in \mathcal{K}_l$, we are interested in calculating the order-$k$ tensor ${\bf Y}_l$, where each element of ${\bf Y}_l$ is the matching score among $k$ nodes. To \textcolor{black}{consider the} node-level embedding vectors from $f_{\text{GNN}}$ \textcolor{black}{to} efficiently \textcolor{black}{calculate} the matching score, we formally define a new scoring function:
\begin{definition}[Prototype scoring function] Given a set of $b$-dimensional vectors $\{{\bf v}_i\}_{i=1}^k$, we define the {\it prototype scoring} function $s: \mathbb{R}^b \times \cdots \times \mathbb{R}^b \rightarrow \mathbb{R}$ as follows:
\begin{equation} \label{scoringfunctiondefinition}
    s({\bf v}_1, \cdots, {\bf v}_k) = {\bf 1}^{T} ({\bf v}_1 \odot \cdots \odot {\bf v}_k),
\end{equation}
where $\odot$ indicates the element-wise product and ${\bf 1}\in\mathbb{R}^b$ is the all-ones vector.
\label{definition:scoringfunction}
\end{definition} 


We now focus on calculating each element of ${\bf Y}_l$ by using the node-level embeddings as input of the prototype scoring function $s$. To this end, we first denote ${\bf X}_l \in \mathbb{R}^{|\mathcal{V}_1| \times \cdots \times |\mathcal{V}_k|}$ as another order-$k$ tensor such that ${\bf X}_l[i_1,\cdots,i_k] \triangleq s({\bf h}_{G_1}^{i_1}, \cdots, {\bf h}_{G_k}^{i_k})$. Each element of ${\bf X}_l$ represents a matching score that encodes the likelihood that $k$ nodes, each of which comes from the corresponding $k$ graphs in the set $\mathcal{K}_l$, match well on the node-level embedding space according to the GNN model $f_\text{GNN}$. In order to normalize ${\bf X}_l$ and encode additional information, we then obtain ${\bf Y}_l$ by going through two operations $u_{k'}^1 (\cdot)$ and $u_{k'}^2 (\cdot)$ on ${\bf X}_l$ for $k'\in\{1,\cdots,k\}$ that modify the values of the input tensor while not changing the dimensions: 1) $u_{k'}^1 ({\bf X}_l)$ applies the softmax normalization along the $k'$-th dimension; and 2) $u_{k'}^2 ({\bf X}_l)$ returns the average matching score along the $k'$-th dimension and is expressed as $\sigma(\sum_{i_{k'}=1}^{|\mathcal{V}_{k'}|} {\bf X}_l [\cdots,i_{k'},\cdots])/|\mathcal{V}_{k'}|$, where $\sigma(\cdot)$ is the sigmoid function. Finally, we are capable of calculating ${\bf Y}_l$ by taking the average of the element-wise products of $u_{k'}^1 ({\bf X}_l)$ and $u_{k'}^2 ({\bf X}_l)$ over all dimensions:
\begin{equation}
    {\bf Y}_l = \dfrac{1}{k}\left(\sum_{k'=1}^{k} u_{k'}^1({\bf X}_l) \odot u_{k'}^2({\bf X}_l)\right).
\end{equation} 

For our matching score calculation, we \textcolor{black}{address the following} computational complexity issues.
\begin{remark}
Although the inner product is a natural way of encoding similarities between two node-level embedding vectors, its extension to the case of $k\ge3$ is not straightforward. Alternative designs of the prototype scoring function include, but \textcolor{black}{are} not limited to\textcolor{black}{,} the approach for taking the average of the inner products between each vector and the arithmetic mean vector. \textcolor{black}{This} approach requires the calculation of all possible inner products between pairs of $k$ embedding vectors, which is \textcolor{black}{inefficient because} such pairwise comparisons take $\mathcal{O}(k^2)$. On the other hand, it is possible to calculate the matching score among $k$ nodes with a linear scaling of $k$ (i.e., $\mathcal{O}(k)$) by using the prototype scoring function $s$ in~(\ref{scoringfunctiondefinition}), which is much more efficient. Additionally, the prototype scoring function based on such an alternative approach does not naturally boil down to the inner product when $k=2$. Moreover, we empirically found that the output of \textsf{PAGE} using such an alternative is shown to be less stable for prototype discovery.
\label{remark:scoringfunction}
\end{remark}

\begin{remark}
\textcolor{black}{To} reduce the search space, we \textcolor{black}{can exclude} node tuples in the matching tensor ${\bf Y}_l$ such that the feature vectors of the corresponding nodes do not coincide. To achieve this, we post-process ${\bf Y}_l$ by assigning zeros to the elements of ${\bf Y}_l$\textcolor{black}{, where the} feature vectors of nodes are not identical to each other. For example, suppose that $k=3$ along with the order-3 matching tensor ${\bf Y}_l$. For each element ${\bf Y}_l [i,j,m]$, we examine the feature vectors of nodes, i.e., ${\bf x}_1^i$, ${\bf x}_2^j$, and ${\bf x}_3^m$. We do not modify the value of ${\bf Y}_l [i,j,m]$ if ${\bf x}_1^i={\bf x}_2^j={\bf x}_3^m$; we assign zeros to ${\bf Y}_l [i,j,m]$ otherwise. Thus, it is possible to avoid searching for such zero-injected node tuples in ${\bf Y}_l$. This post-processing does not deteriorate the performance of \textsf{PAGE} while substantially reducing the computational complexity of prototype search.

\end{remark}

Next, we create $k$ empty node lists $\mathcal{S}_1, \cdots, \mathcal{S}_k$ in which node indices will be included during the iterative search. As illustrated in the top-left of Fig.~\ref{figure:mainiteration}, for $k=3$, three empty lists $\mathcal{S}_1$, $\mathcal{S}_2$, and $\mathcal{S}_3$ are being initialized. The node indices in each list after the final iteration correspond to the nodes in the prototype graph $g_{l}^{(c)}$. 

\subsubsection{Core Prototype Search Module}~\label{subsubsection:mainproto}
For the remaining part of our prototype discovery phase, we would like to materialize the prototype discovery function $q_{\text{proto}}$, which is designed to extract common subgraph patterns in a reliable manner. In a nutshell, as depicted in Fig.~\ref{endtoendPAGE}, we run multiple search sessions, each producing one prototype graph candidate. After running all search sessions, we feed each prototype candidate to the underlying model $f_{\text{GNN}}$, and acquire the one with the highest output probability $p_\text{GNN}$ as the final prototype graph. This ensures that we leverage various subgraphs as candidates for the final prototype graph, thereby increasing the chances of retrieving a trustworthy prototype graph that conforms to the underlying GNN model.

More concretely, we introduce a search budget, denoted as the variable $budget$, which determines the number of search sessions that we run. Each search session is composed of iterative subgraph search (Step 2) and prototype retrieval (Step 3). In other words, for a given {\it budget}, we run up to {\it budget} sessions for prototype search. When the prototype search is over for all sessions, we select the one having the highest output probability $p_\text{GNN}$ among the $budget$ discovered subgraphs as the final prototype graph $g_l^{(c)}$ for the target class $c$.

\subsubsection{Iterative Subgraph Search (Step 2)}~\label{subsubsection:iterativesubgraphsearch}
We now describe the iterative subgraph search step in each search session. In this step, we iteratively find $k$-tuples of node indices to be inserted into the created $k$ node lists. Each iteration starts by collecting all possible $k$-tuples to be selected. In Iteration 0 (i.e., initial search), all combinations of $k$ nodes across $k$ different graphs in the set $\mathcal{K}_l$ are valid candidates for selection. For the rest of the iterative search process, we construct node sequences by traversing for each graph; that is, combinations only from neighbors for each of the $k$ selected nodes in the previous iteration are valid candidates for selection (See the block of Iteration 1 in Fig.~\ref{figure:mainiteration} for $k=3$). Selection among the valid candidates for the $k$-tuples is determined based on the corresponding value in ${\bf Y}_l$. Each $k$-tuple with the highest matching score is chosen out of possible combinations from the sets of neighbors of already selected nodes during the iteration process, except the initial search. For the initial search, the $k$-tuple with the $b$-th highest matching score is selected, where $b\in\{1,\cdots,budget\}$ indicates the index of the search session described in Section~\ref{subsubsection:mainproto}. We describe this step more concretely when $k=3$ as follows.

\begin{example} \label{example:mainiteration}
As illustrated in Fig.~\ref{figure:mainiteration}, in our example, suppose that $k=3$ and node indices 6, 5, and 5 are in the node lists $\mathcal{S}_1$, $\mathcal{S}_2$, $\mathcal{S}_3$, respectively, in Iteration 0 (initialization). In Iteration 1, we focus only on the set of neighbors for each node, which are $\mathcal{N}_{G_1}(v_1^6) = \{4, 5\}$, $\mathcal{N}_{G_2}(v_2^5) = \{3, 4\}$, and $\mathcal{N}_{G_3}(v_3^5) = \{3\}$, where $\mathcal{N}_{G_i}(v)$ is the set of neighbors of node $v$ in $G_i$. Then, all possible candidates are the Cartesian product of the three sets of neighbors: $\mathcal{N}_{G_1}(v_1^6) \times \mathcal{N}_{G_2}(v_2^5) \times \mathcal{N}_{G_3}(v_3^5) = \{4, 5\} \times \{3, 4\} \times \{3\}$. In consequence, the possible node candidates are given by $(4,3,3)$, $(4,4,3)$, $(5,3,3)$, and $(5,4,3)$ as depicted in Iteration 1.
\end{example}

Each node index of the selected $k$-tuple is added to each node list $\mathcal{S}_1, \cdots, \mathcal{S}_k$. After selection, we update ${\bf Y}_l$ by multiplying a certain decay factor along each dimension from the selected $k$-tuple.\footnote{We have empirically found that such an update of ${\bf Y}_l$ with a decay factor enables us to avoid searching for the previously selected $k$-tuples, resulting in a more stable subgraph search process.} The whole subgraph search iteration terminates when either the allowed maximum number of iterations is reached or the size of the resultant subgraph exceeds a pre-defined value.

\subsubsection{Prototype Retrieval (Step 3)}~\label{subsubsection:prototyperetrieval}
As the next step, we are capable of discovering $k$ subgraphs, each of which is retrieved from the node list $\mathcal{S}_i$ for $G_i$. In Step 3 of prototype discovery for each search session, we describe how to retrieve a candidate of the final prototype graph by evaluating the output probabilities $p_\text{GNN}$ for each discovered subgraph from the list $\mathcal{S}_i$. Specifically, we feed each subgraph as input to the model $f_{\text{GNN}}$ to get $p_\text{GNN}$ and choose the subgraph with the highest $p_\text{GNN}$ among $k$ subgraphs (see the bottom-right in Fig.~\ref{figure:mainiteration} for $k=3$).

\textcolor{black}{As} we run {\it budget} search sessions, we acquire {\it budget} prototype candidates. We retrieve the final prototype graph $g_l^{(c)}$ by selecting the prototype candidate \textcolor{black}{with} the highest output probability $p_\text{GNN}$ among {\it budget} candidates.

\section{EXPERIMENTAL EVALUATION} \label{section:experiment}

In this section, we first describe synthetic and real-world datasets used in the evaluation. We also present the benchmark GNN explanation method for comparison. After describing our experimental settings, we comprehensively evaluate the performance of \textsf{PAGE} and benchmark methods. The source code for \textsf{PAGE} is made publicly available online.\footnote{github.com/jordan7186/PAGE.}

\subsection{Datasets} \label{subsection:datasets}
In our study, two synthetic datasets, including the BA-house and BA-grid datasets, and \textcolor{black}{four} real-world  datasets, including the Solubility, MUTAG, Benzene, \textcolor{black}{and MNIST-sp} datasets, are used for the evaluation of our proposed \textsf{PAGE} method. The ground truth explanations (i.e., ground truth prototypes) as well as the ground truth labels for graph classification are available for \textcolor{black}{all} datasets. The main statistics of each dataset are summarized in Table~\ref{table:experiment_statistics}. We describe important characteristics of the datasets.

\begin{table}[t]
\centering
  \begin{tabular}{lcccc}
        \toprule
        \textbf{Dataset} & $n$ & $\sum_{i} \mathcal{V}_i$ (Avg.) & $\sum_{i} \mathcal{E}_i$ (Avg.) & TP \\
        \midrule
        BA-house & 2,000 & 21,029 (10.51) & 62,870 (31.44) & 1.000\\
        BA-grid & 2,000 & 29,115 (14.56) & 91,224 (45.61) & 0.9583\\
        \midrule
        Benzene & 12,000 & 246,993 (20.58) & 523,842 (43.65) & 0.9444\\
        MUTAG & 4,337 & 131,488 (30.32) & 266,894 (61.54) & 0.7247\\
        Solubility & 708 & 9,445 (13.34) & 9,735 (13.75) & 0.8717\\
        MNIST-sp & 70,000 & 5,250,000 (75) & 41,798,306 (696.63) & 0.7595\\
        \bottomrule
  \end{tabular}
  \caption{Statistics of the six datasets and test performance of the GNN model trained for each dataset, where $n$, $\sum_i \mathcal{V}_i$, $\sum_i \mathcal{E}_i$, and TP denote the number of graphs, the total number of nodes, the total number of edges, and the test performance of the GNN model, respectively. We also report the average number of nodes and edges per graph for each dataset in the parenthesis.}
  \label{table:experiment_statistics}
\end{table}

{\bf BA-house} and {\bf BA-grid}. Inspired by~\cite{ying2019gnnexplainer}, we generate new synthetic datasets \textcolor{black}{for graph classification}. In both datasets, we use the Barab\'{a}si-Albert (BA) model as a backbone graph, and complete each graph by attaching certain subgraph patterns (or motifs), which serve as the ground truth explanations. Both datasets contain two classes, where one class includes the ground truth explanation(s) in the underlying graph, while another class has an incomplete subgraph pattern. To generate the dataset, we first assign the class label 0 or 1 to a graph to be generated. We then generate a backbone graph using the BA model consisting of $5$ to $10$ nodes. If the assigned class label is $0$, then we connect additional subgraph patterns to the backbone graph. Otherwise, we connect an incomplete subgraph pattern to the backbone graph. For BA-house, the subgraph pattern is a house-shaped subgraph, while, for BA-grid, a grid-like subgraph is used. In the case of BA-house, either single {\em or} double house-shaped subgraphs are randomly added within Class $0$. Each node has a one-hot encoded feature vector indicating whether the node belongs to the backbone graph, the head of the motif (i.e., the ground truth explanation), or the body of the motif, depending on the node's position in the given graph.\\

{\bf Benzene}~\cite{sanchez2020gnneval}. The Benzene dataset contains molecules, where nodes and edges represent atoms and chemical bonds, respectively. The graphs are partitioned (labeled) into two different classes with respect to the existence of the benzene structure within the molecule (i.e., the structure having a six-carbon ring as the dataset neglects the Hydrogen atom). Each node has the corresponding one-hot encoded feature vector representing one possible atom type, which can be one of Carbon, Nitrogen, Oxygen, Fluorine, Iodine, Chlorine, and Bromine.\\

{\bf MUTAG}~\cite{debnath1991mutag}. The MUTAG dataset contains graphs representing molecules, where nodes represent different atoms, similarly as in the Benzene dataset. The graphs are partitioned into two different classes according to their mutagenic effect on a bacterium~\cite{ying2019gnnexplainer}. The node features are one-hot encoded vectors that represent types of atoms, including Carbon, Nitrogen, Oxygen, Fluorine, Iodine, Chlorine, Bromine, and Hydrogen. In our experiments, the edge labels are ignored for simplicity.\\

{\bf Solubility}~\cite{baldassarre2019infectionsolubility}. The Solubility dataset is composed of real-world molecules with different levels of solubility, where nodes and edges represent atoms and their chemical bonds, respectively. In our experiments, we ignore the edge connection types. Although this dataset was originally intended to be used as a regression task, it is partitioned into two classes for the graph classification task. We label molecules with log solubility values lower than $-4$ as $0$, and those with values higher than $-2$ as $1$. Rigid carbon structures are considered as the ground truth explanation for insolubility, while $R$-$OH$ chemical groups are treated as the ground truth explanation for solubility. The one-hot encoded feature vector of each node represents types of atoms, which can be one of Carbon, Nitrogen, Oxygen, Fluorine, Iodine, Chlorine, Phosphorus, Sulfur, Hydrogen, and Bromine.\\

{\bf MNIST-sp}~\cite{monti2017mnistsp}. MNIST-sp is a dataset for graph classification originated from the MNIST dataset used for image classification. Since local pixels are processed to form a node (superpixel) for each image, the image classification is transformed into a graph classification problem. Each graph contains 75 nodes, and edges are connected depending on whether the superpixels are neighbors in the original image. The node features include the center positions of the superpixel and the \textcolor{black}{continuous} values of the pixels. Therefore, we \textcolor{black}{categorize the continuous values to discretize them}. The pixel values are categorized into three groups according to their original values: 0 (completely black), $(0, 0.5]$ (dark), $(0.5, 1]$ (bright). The position values are categorized to an evenly divided $14 \times 14$ grid of the image via their association.

Note that model-level explanation methods aim to \textcolor{black}{identify explanations based} on the classes of each dataset. Thus, in all the experiments, model-level explanation methods use the class including a precise subgraph pattern as input and then attempt to retrieve the class-distinctive subgraph as the prototype graph.

\subsection{Benchmark Method}
In this subsection, we present a state-of-the-art method for comparison. \textcolor{black}{As the representative model-level explanation, we use XGNN~\cite{yuan2020xgnn} as our benchmark method,} which trains a graph generator that generates a subgraph pattern (i.e., a prototype graph) via reinforcement learning in the sense of maximizing a certain prediction of the underlying GNN model. \color{black}More specifically, the graph generator is trained to add edges in an iterative fashion. At step $t$ of the generation, the generated graph $G_{t+1}$ is fed into the underlying GNN to calculate the reward function $R_t$:
\begin{equation}
    R_t=R_{t,f_{\text{GNN}}}(G_{t+1}) + \lambda_1 \dfrac{\sum_{i=1}^m \text{Rollout}(G_{t+1})}{m} + \lambda_2 R_{t, r},
\end{equation}
where $\lambda_1$ and $\lambda_2$ are hyperparameters; $R_{t,f_{\text{GNN}}}(G_{t+1})$ is the predicted class probability of the underlying GNN $f_{\text{GNN}}$; $\text{Rollout}(\cdot)$ is the rollout operation~\cite{Yu2017rollout}; $m$ is the number of times rollout is performed, and $R_{t, r}$ is the additional term that provides domain-specific rewards based on various graph rules.\color{black}


\begin{figure*}[t]
\centering
\includegraphics[width=0.9\textwidth]{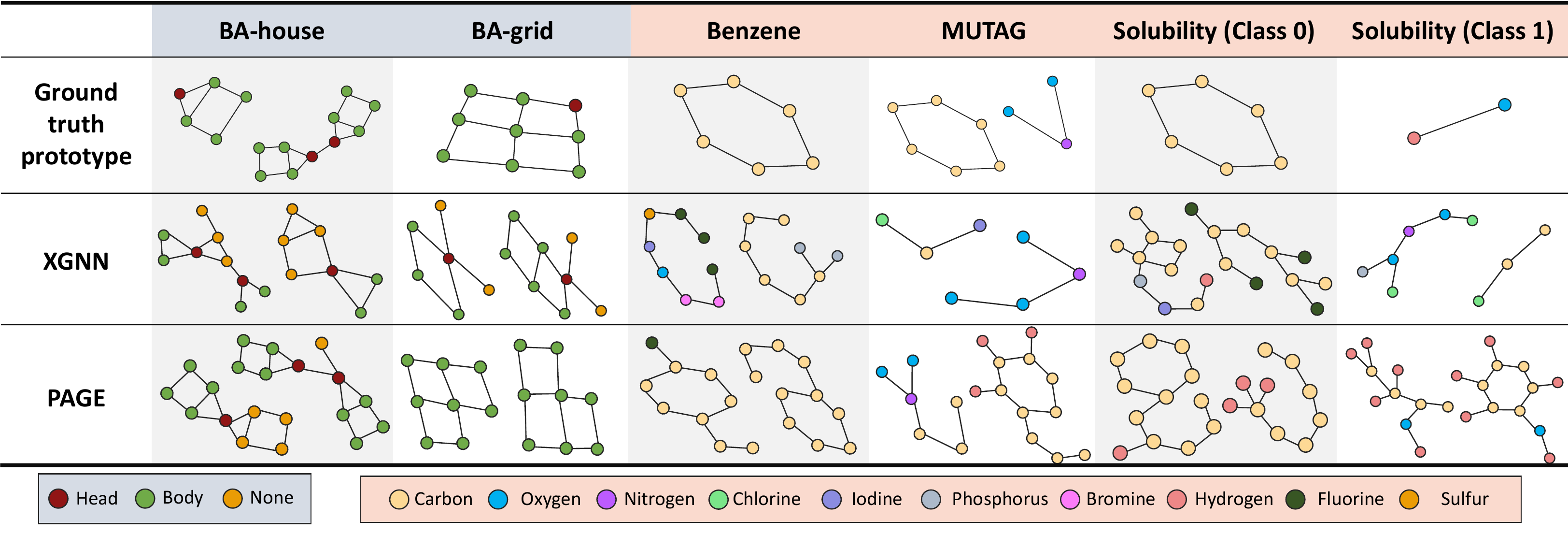}
\caption{Qualitative comparison of \textsf{PAGE} and XGNN for five datasets (excluding MNIST-sp, which is separately presented in Fig.~\ref{fig:MNIST_sp_results}), where each node is colored differently according to the types of nodes.}
\label{figure:qual0}
\end{figure*}

\subsection{Experimental Settings}
We first describe the settings of the GNN model. We adopt GCN~\cite{kipf2017gcn} as one of the widely used GNN architectures. The summation function is used as a readout function. We use 2 GNN layers for the BA-house and BA-grid datasets and 3 GNN layers for other three datasets unless otherwise stated. We set the dimension of each hidden latent space to 32 for all the datasets. We train the GNN model with Adam optimizer~\cite{kingma2015adam} with a learning rate of 0.001 and a batch size of 16. For all five datasets \textcolor{black}{excluding MNIST-sp}, we split each dataset into training/validation/test sets with a ratio of 90/5/5\%. \textcolor{black}{For the MNIST-sp dataset, we follow the given train/test split and further split the training set into train/val sets with a ratio of 5:1.} The training set is used to learn the model parameters with the cross-entropy loss; and the validation set is used for early stopping with a patience of 5 epochs. \textcolor{black}{Unless otherwise specified, we use all the graphs in the training set to run \textsf{PAGE}.} We report the graph classification performance on the test set along with the datasets in Table~\ref{table:experiment_statistics}.

Next, we turn to describing the settings of explanation methods including \textsf{PAGE} and XGNN. In \textsf{PAGE}, the implementations of clustering with the GMM are those from the scikit-learn Python package~\cite{pedregosa2011scikitlearn}. \textcolor{black}{We set $L=2$ and $k=3$ since we have found that such a setting produces stable results across different datasets.} We also set $budget$, indicating the number of search sessions, to 5. Additionally, we set the decay rate and the maximum number of iterations used in the iterative subgraph search (Step 2 of Phase 2) as 10 and 1,000, respectively. For the implementations of XGNN, we tuned the hyperparameters by essentially following the same settings as those in the original paper~\cite{yuan2020xgnn}. 

The GNN model and explanation methods used in our experiments were implemented with Python 3.8.12, PyTorch 1.11.0, PyTorch Geometric 2.0.4~\cite{fey2019pyg}, and Captum 0.5.0~\cite{kokhlikyan2020captum}, and were run on a machine with an Intel Core i7-9700K 3.60 GHz CPU with 32GB RAM and a single NVIDIA GeForce RTX 3080 GPU.

\subsection{Experimental Results}

Our experiments are designed to answer the following eight key research questions (RQs).
\begin{itemize}
    \item {\bf RQ1:} How are the model-level explanations of \textsf{PAGE} {\it qualitatively} evaluated in comparison with the benchmark method?
    \item {\bf RQ2:} How are the model-level explanations of \textsf{PAGE} {\it quantitatively} evaluated in comparison with the benchmark method?
    \item {\bf RQ3:} When does \textsf{PAGE} return the final prototype graph during multiple search sessions?
    \item {\bf RQ4:} How do the explanations of \textsf{PAGE} relate to instance-level explanation methods?
    \item {\bf RQ5:} How many input graphs does \textsf{PAGE} require for model-level explanations? 
    \item {\bf RQ6:} How effective is the prototype scoring function in \textsf{PAGE} in comparison with na\"ive alternatives?
    \item {\bf RQ7:} How expensive is the computational complexity of \textsf{PAGE} in comparison with the benchmark method?
    \item \color{black}{\bf RQ8:} How does \textsf{PAGE} perform for datasets without explicit ground truth explanations?\color{black}
\end{itemize}

\subsubsection{Qualitative Analysis of Model-Level Explanation Methods (RQ1)}

\begin{table*}[!ht]
  \centering
  \subfloat[\textcolor{black}{Accuracy (↑).}]{%
  \resizebox{.48\textwidth}{!}{%
  \begin{tabular}{lccccccc}
        \toprule
        \textbf{Method} & BA-house & BA-grid & Solubility & MUTAG & Benzene & MNIST-sp\\
        \midrule
        \textsf{PAGE}  & 0.5238 & 0.8571 & 0.3290 & 0.9090 & 0.6667 & N/A \\
        XGNN  & 0.2500 & 0.3200 & 0.2341 & 0.6875 & 0.2500  & N/A \\
        \bottomrule
  \end{tabular}
  }
}
  \subfloat[\textcolor{black}{Density (↓).}]{%
  \resizebox{.48\textwidth}{!}{%
  \begin{tabular}{lccccccc}
        \toprule
        \textbf{Method} & BA-house & BA-grid & Solubility & MUTAG & Benzene & MNIST-sp \\
        \midrule
        \textsf{PAGE}  & 0.1481 & 0.1563 & 0.0462 & 0.1389 & 0.1111 & 0.2195 \\
        XGNN  & 0.1235 & 0.1667 & 0.1195 & 0.1875  & 0.1094 & 0.3593 \\
        \bottomrule
  \end{tabular}
  }
  }

  \subfloat[Consistency (↓).]{%
  \resizebox{.48\textwidth}{!}{%
  \begin{tabular}{lcccccc}
        \toprule
        \textbf{Method} & BA-house & BA-grid & Solubility & MUTAG & Benzene & MNIST-sp \\
        \midrule
        \textsf{PAGE}  & 0.0308 & 0.0615 & 0.0846 & 0.1216 & 0.0639 & 0.1025 \\
        XGNN  & 0.2152 & 0.2705 & 0.3213 & 0.1269  & 0.2227 & 0.0242 \\
        \bottomrule
  \end{tabular}
  }
  }
  \subfloat[Faithfulness (↑).]{%
  \resizebox{.48\textwidth}{!}{%
  \begin{tabular}{lcccccc}
        \toprule
        \textbf{Method} & BA-house & BA-grid & Solubility & MUTAG & Benzene & MNIST-sp \\
        \midrule
        \textsf{PAGE}  & 0.7340 & 0.5636 & 0.2164 & 0.4430 & 0.2364 & 0.8182 \\
        XGNN  & -0.4037 & -0.1636 & 0.0983 & 0.2504  & -0.3091 & 0.1273 \\
        \bottomrule
  \end{tabular}
  }
  }
\caption{\textcolor{black}{Quantitative assessment results for \textsf{PAGE} and XGNN for six datasets. Here, the accuracy \textcolor{black}{(the higher the better)} compares the edge correspondences to the ground truth explanations; the density \textcolor{black}{(the lower the better)} measures the ratio of edges over nodes; the consistency \textcolor{black}{(the lower the better)} measures the variability of the output probabilities for the explanations over different settings of GNN models; and the faithfulness \textcolor{black}{(the higher the better)} measures the correlation between the output probabilities of explanations and the GNN's test accuracies.}}\label{table:quantitativeeval}
\end{table*}

We perform a qualitative evaluation of both \textsf{PAGE} and XGNN by visualizing their explanations (i.e., prototype graphs) along with the ground truth explanations for each dataset. As illustrated in Fig.~\ref{figure:qual0}, we observe that the explanations discovered by \textsf{PAGE} and the ground truth explanations exhibit a high similarity in terms of the graph structure and node features (i.e., types of nodes), while XGNN mostly fails to produce prototypes that contain the entire ground truth explanations. From Fig.~\ref{figure:qual0} and Fig.~\ref{fig:MNIST_sp_results}, our findings for each dataset are as follows:
\begin{itemize}
    \item For the BA-house dataset, \textsf{PAGE} successfully returns the class-discriminative subgraphs that we add to the backbone graph, that is, the house-shaped subgraphs on BA-house. However, the prototypes generated by XGNN contain only parts of the house-shaped subgraphs. For instance, the outcome from XGNN contains only three out of five nodes of the house-shaped subgraph, thus making the explanation incomplete. On the other hand, \textsf{PAGE} can produce much more precise explanations by returning two types of prototype graphs containing either a single or double house-shaped subgraph. Such preciseness of \textsf{PAGE} is possible primarily due to the usage of clustering of graph-level embeddings.
    \item The results from the BA-grid dataset show a tendency similar to those on BA-house. \textsf{PAGE} returns most of grid-like subgraphs, while XGNN recovers smaller portions of them.
    \item For the case of the Benzene dataset, \textsf{PAGE} correctly identifies the ring structure with 6 carbon atoms as a part of the prototype graph, while XGNN does not generate any carbon ring. Although XGNN produces six carbon atoms, it fails to connect the atoms to a closed ring structure. 
    \item For the MUTAG dataset having two types of prototype graphs, \textsf{PAGE} identifies both carbon rings and $NO_2$ chemical groups. However, XGNN only recovers $NO_2$ while generating small molecules. 
    \begin{figure}[t]
         \centering
         \begin{subfigure}[b]{0.43\columnwidth}
             \centering
             \includegraphics[width=\columnwidth]{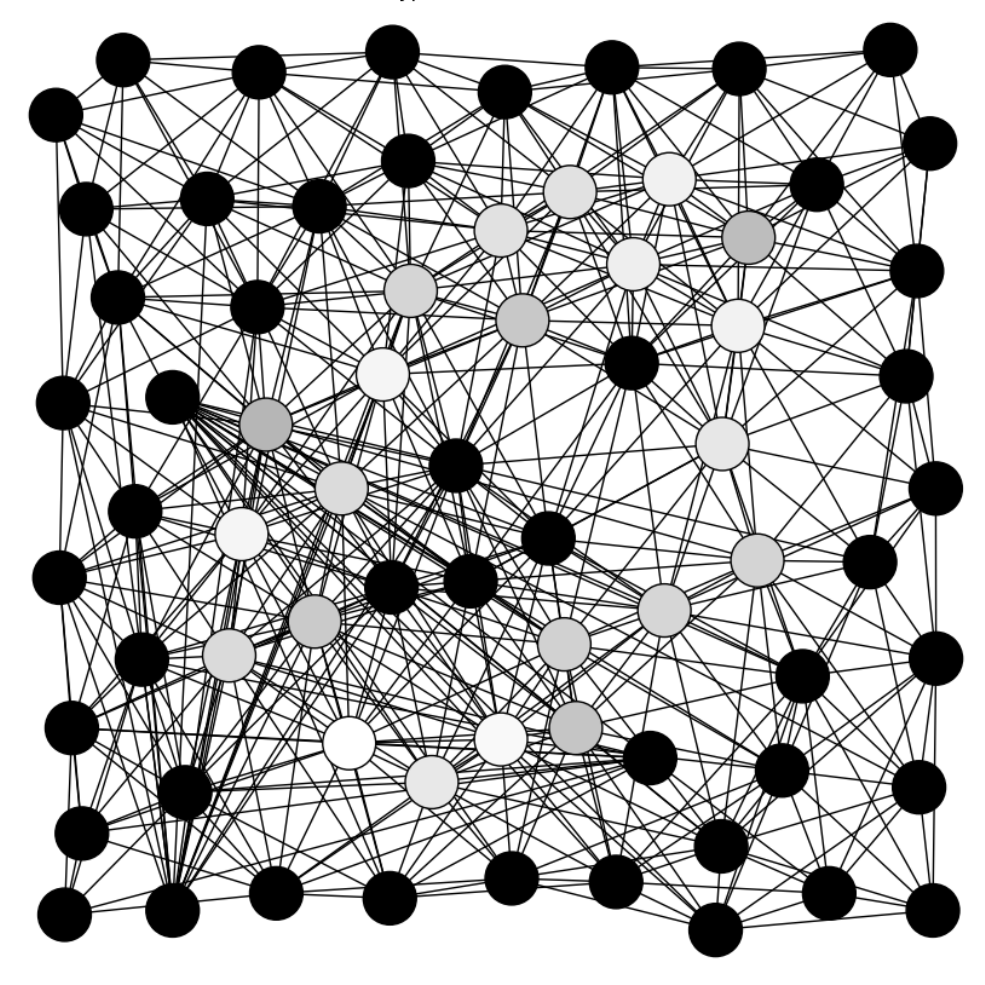}
             \caption{\textsf{PAGE}.}
             \label{fig:MNIST_sp_result1}
         \end{subfigure}
         \hfill
         \begin{subfigure}[b]{0.43\columnwidth}
             \centering
             \includegraphics[width=\columnwidth]{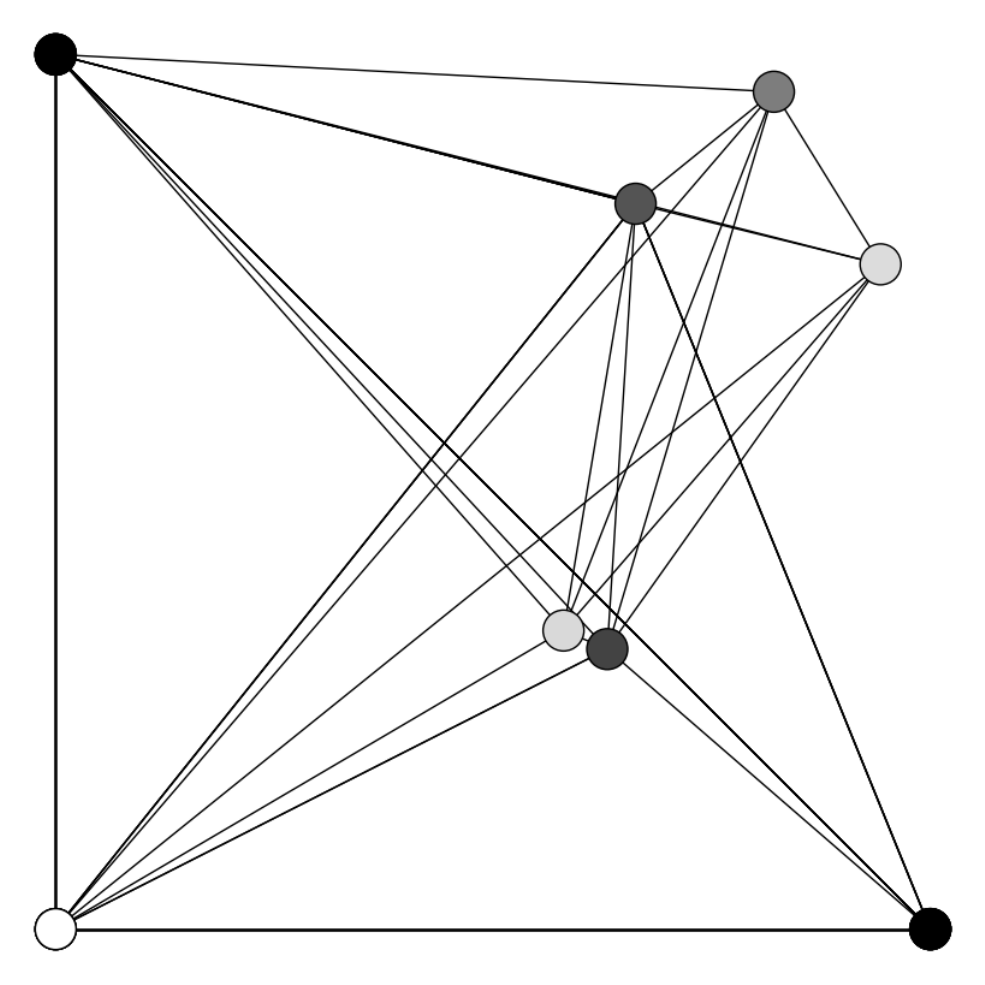}
             \caption{XGNN.}
             \label{fig:MNIST_sp_result2}
         \end{subfigure}
            \caption{Qualitative comparison of \textsf{PAGE} and XGNN for Class 0 on the MNIST-sp dataset.}
            \label{fig:MNIST_sp_results}
    \end{figure}
    \item We analyze model-level explanations for two classes on Solubility. For Class 0 (i.e., the case classified as insoluble molecules), \textsf{PAGE} tends to return carbon atoms and occasionally adds hydrogen atoms; XGNN also produces prototypes with mostly carbon atoms but contains a large portion of non-carbon atoms. Unlike the case of \textsf{PAGE}, XGNN fails to produce the carbon ring structure. For Class 1 (i.e., the case classified as soluble molecules), \textsf{PAGE} correctly identifies the $R$-$OH$ chemical group for all prototypes; in contrast, XGNN fails to include $R$-$OH$ for its explanations.
    \item We finally analyze the results for Class 0 on MNIST-sp. As depicted in Fig.~\ref{fig:MNIST_sp_results}, the number 0 corresponding to the class of interest is visible in the explanation for \textsf{PAGE}, whereas XGNN struggles to generate a sufficient number of new nodes to provide a comprehensive explanation for Class 0. This indicates the benefit of \textsf{PAGE} that discovers the prototype graph in the dataset over its counterpart generating it.
\end{itemize}

\subsubsection{Quantitative Analysis of Model-Level Explanation Methods (RQ2)} \label{subsubsection:RQ2}
For the quantitative analysis, we adopt \textcolor{black}{four} performance metrics: \textcolor{black}{\textit{accuracy}, \textit{density}}, {\it consistency} and {\it faithfulness}.

\begin{figure*}[!ht]
\centering
\begin{subfigure}{.16\textwidth}
    \centering
    \includegraphics[width=\linewidth]{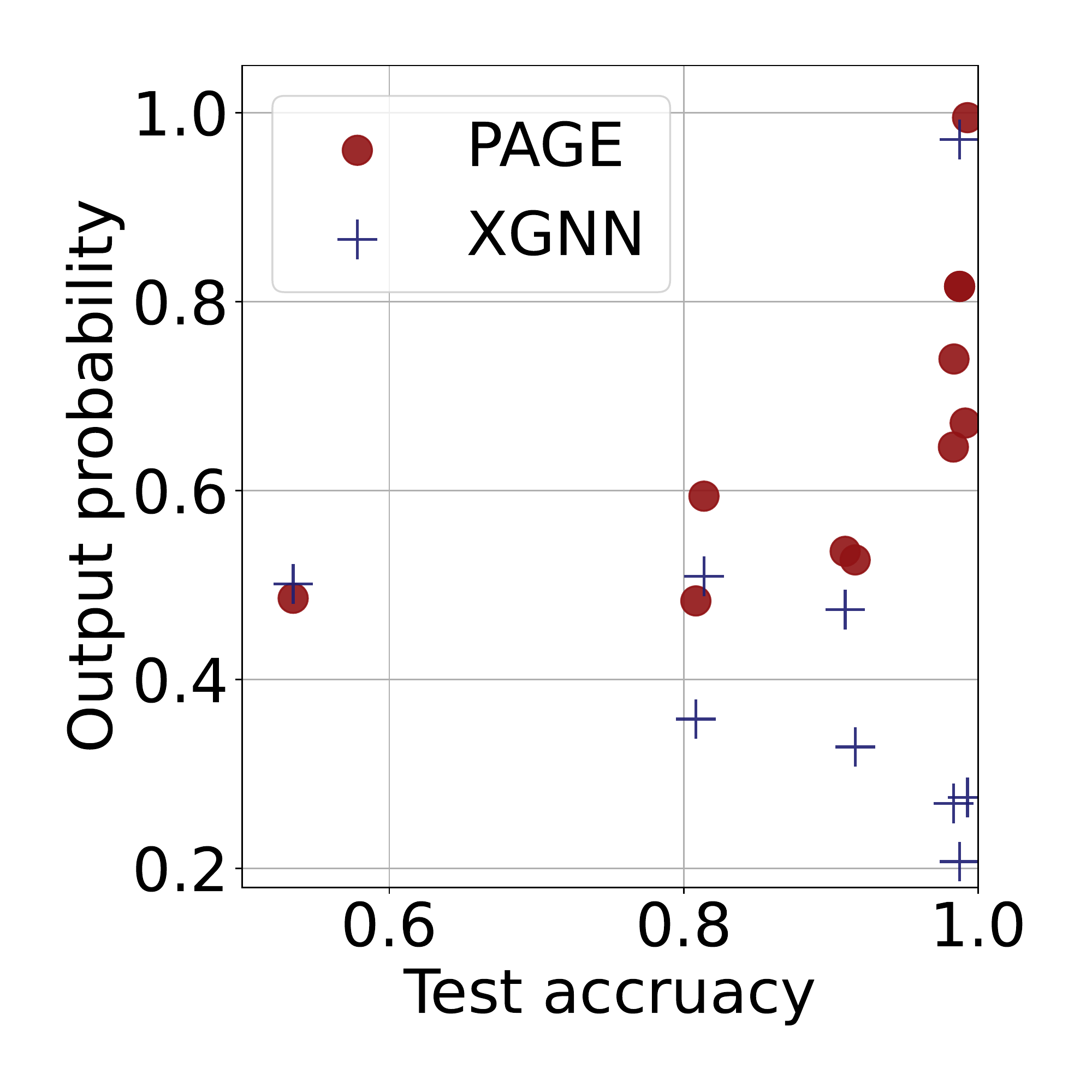}  
    \caption{BA-house.}
    \label{subfig:faithfulbahouse}
\end{subfigure}
\begin{subfigure}{.16\textwidth}
    \centering
    \includegraphics[width=\linewidth]{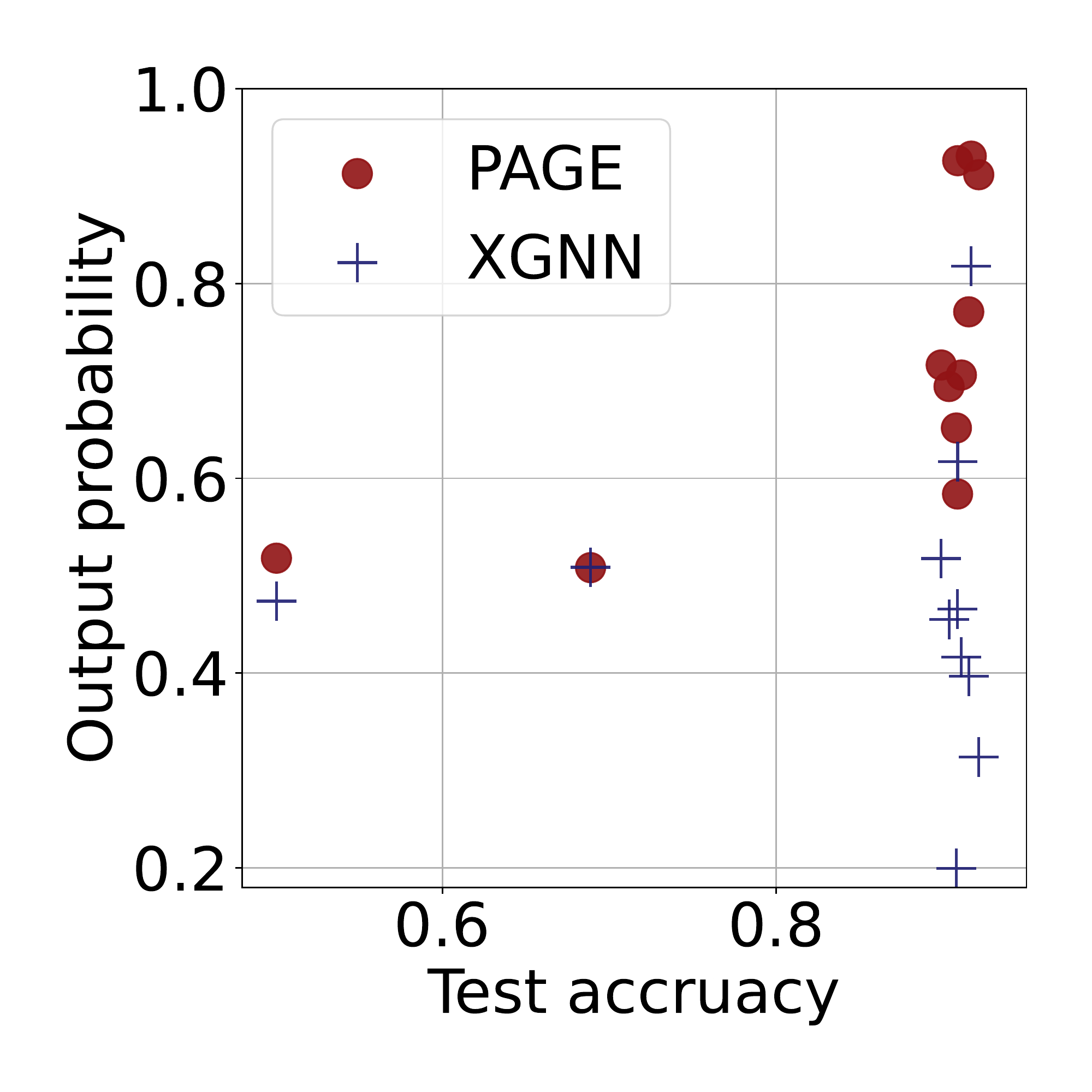}  
    \caption{BA-grid.}
    \label{subfig:faithfulbagrid}
\end{subfigure}
\begin{subfigure}{.16\textwidth}
    \centering
    \includegraphics[width=\linewidth]{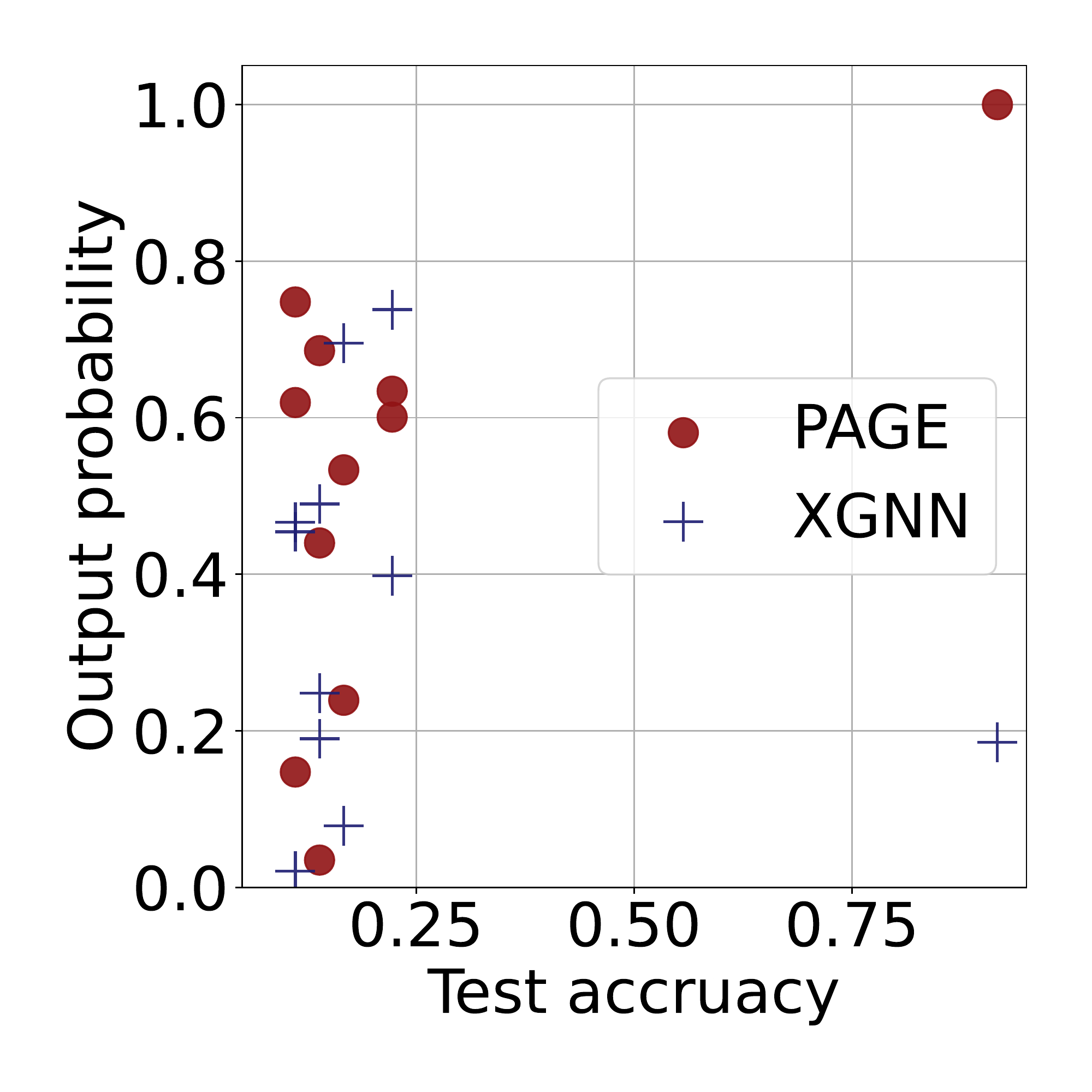} 
    \caption{Solubility.}
    \label{subfig:faithfulsolubility}
\end{subfigure}
\begin{subfigure}{.16\textwidth}
    \centering
    \includegraphics[width=\linewidth]{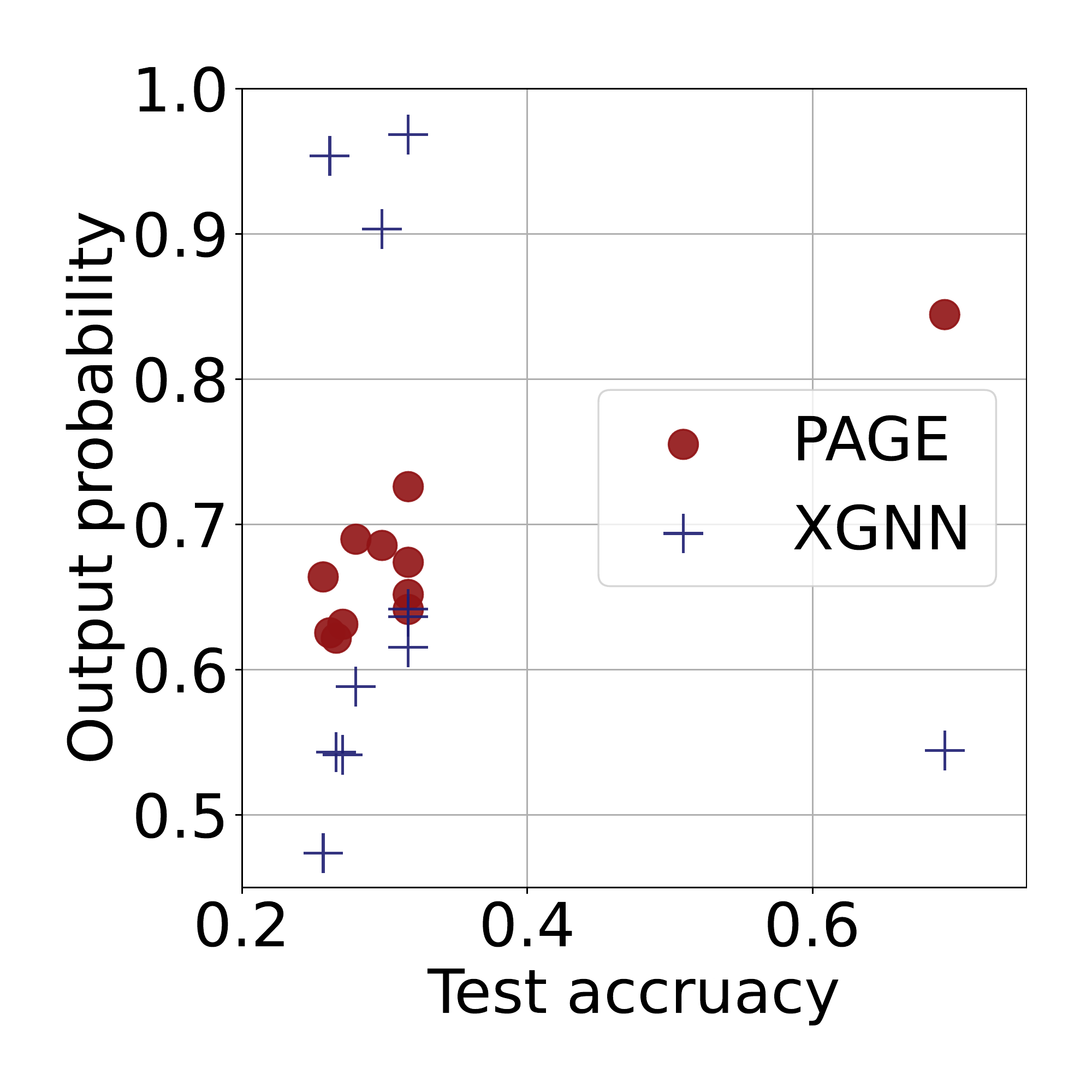}  
    \caption{MUTAG.}
    \label{subfig:faithfulmutag}
\end{subfigure}
\begin{subfigure}{.16\textwidth}
    \centering
    \includegraphics[width=\linewidth]{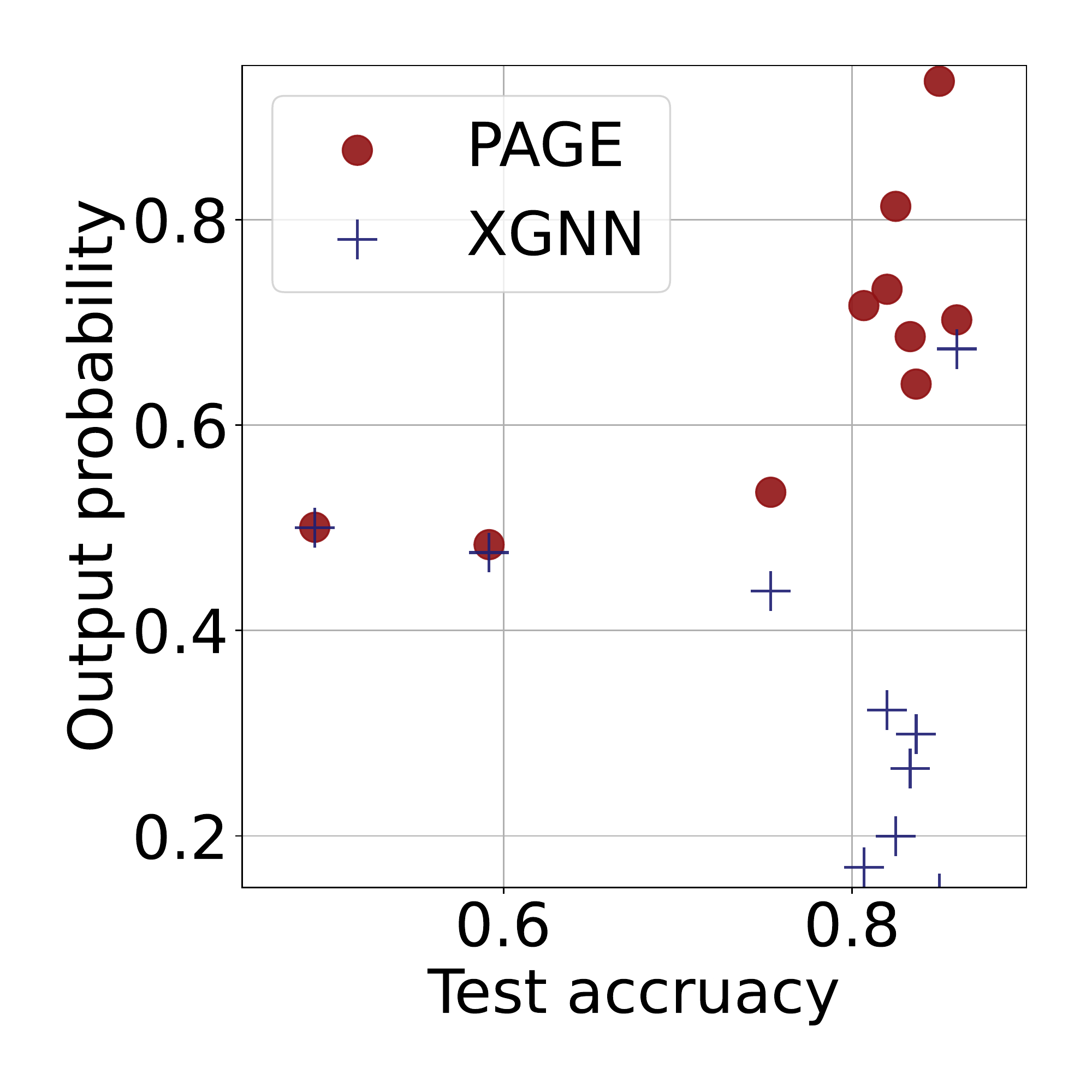}  
    \caption{Benzene.}
    \label{subfig:faithfulbenzene}
\end{subfigure}
\begin{subfigure}{.16\textwidth}
    \centering
    \includegraphics[width=\linewidth]{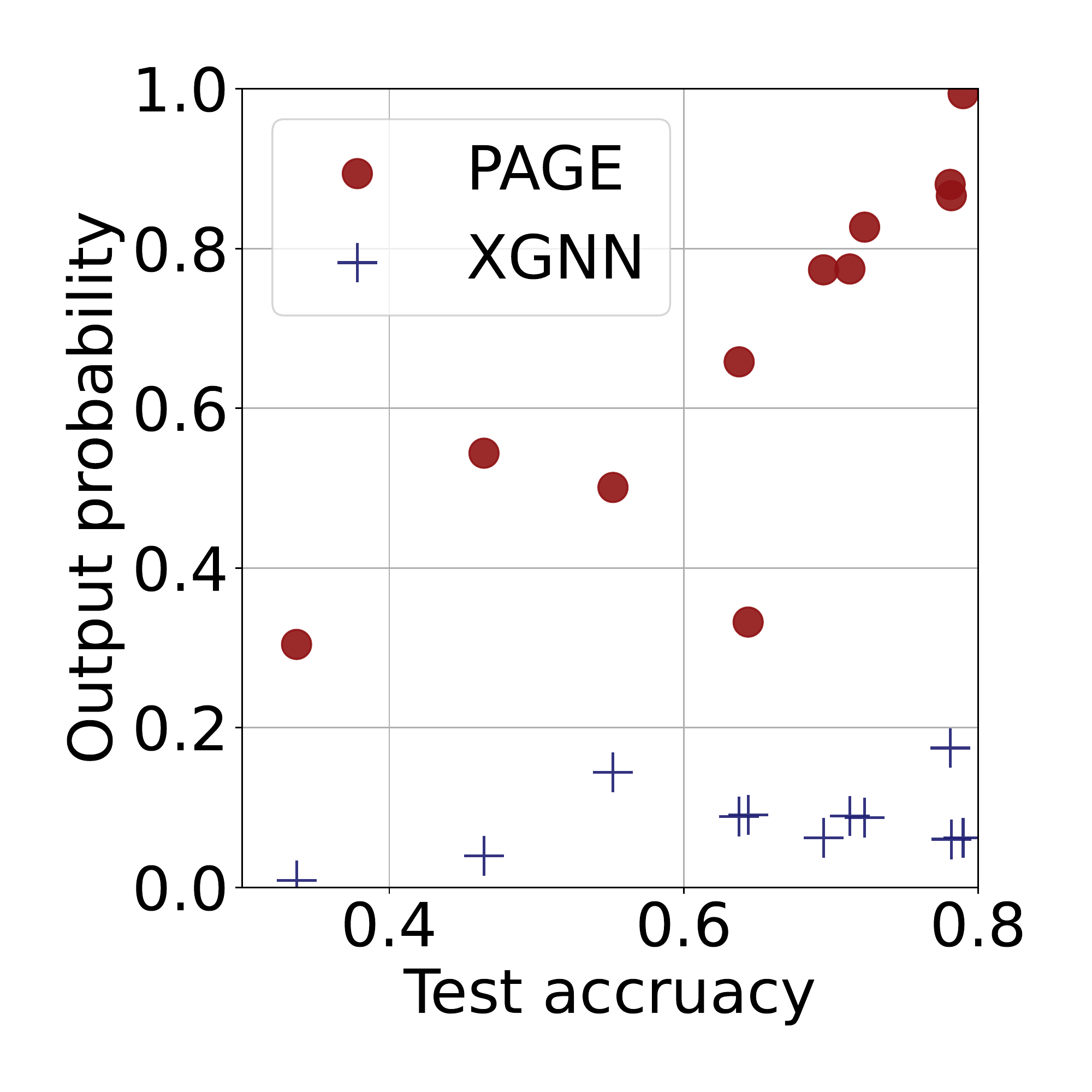}  
    \caption{MNIST-sp.}
    \label{subfig:faithfulmnistsp}
\end{subfigure}
\caption{Visualization of the output probabilities of explanations versus the GNN's test accuracies for measuring the faithfulness when \textsf{PAGE} and XGNN are used as model-level explanation methods.}
\label{figure:quantfaith}
\end{figure*}

\begin{itemize}
\color{black}
    \item \textbf{Accuracy} compares the prototype graph $g$ with the ground truth explanation $g_T$. To this end, we assume that we can acquire node/edge correspondences connecting $g_T$ and $g$ (i.e., node matching between the two graphs) while counting the number of edges in $g$ that belong to a part of $g_T$, which would result in the best explanation performance. In our evaluation, we define the accuracy (Acc.) of model-level explanations as follows: 
    \begin{equation}
    \label{eq:quantitativemetric}
        \text{Acc.} = \dfrac{\text{TP}(g, g_T)}{\text{TP}(g, g_T) + \text{FP}(g, g_T) + \text{FN}(g, g_T)},
    \end{equation}
    where $\text{TP}(g, g_T)$, $\text{FP}(g, g_T)$, and $\text{FN}(g, g_T)$ represent the true positive(s), false positive(s), and false negative(s), respectively, in terms of counting the numbers of relevant nodes and edges between $g$ and $g_T$.\footnote{\textcolor{black}{Note that the accuracy cannot be measured for the MNIST-sp dataset as the dataset does not contain a fixed ground truth explanation $g_T$.}}
    \item \textbf{Density} measures the ratio of the number of edges relative to the square of the number of nodes in the prototype graph $g$.\color{black}
    \item {\bf Consistency}~\cite{sanchez2020gnneval} measures the robustness of explanations for different settings of GNN models. \textcolor{black}{Intuitively,} reliable explanation methods should produce {\it stable} explanations despite different model configurations \textcolor{black}{for the same task}. In our experiments, we calculate the {\it standard deviation} of output probabilities for the explanations (e.g., the prototype graphs retrieved by \textsf{PAGE}) across different GNN hyperparameter settings. To this end, we generate multiple GNN models with various combinations of the hidden layer dimension. For the BA-house and BA-grid datasets, we produce 25 models with $\{4,8,16,32,64\} \times \{4,8,16,32,64\}$ hidden dimension configurations. For the Benzene, Solubility, MUTAG \textcolor{black}{, and MNIST-sp} datasets, we produce 64 models with $\{4,8,16,32\} \times \{4,8,16,32\} \times \{4,8,16,32\}$ hidden dimension configurations. The lower the value of consistency, the better the performance.
    \item {\bf Faithfulness}~\cite{sanchez2020gnneval} measures how closely the explanation method reflects the GNN model's behavior. Following~\cite{sanchez2020gnneval}, we regard the test accuracy of the model as the behavior of interest, and empirically observe the relationship between the output probabilities of explanations and the GNN's test accuracies. In our experiments, faithfulness is measured by the Kendall's tau coefficient~\cite{kendall1938kentalltau} between the output probability $p_\text{GNN}$ for the explanation and the GNN's test accuracy (i.e., the performance on graph classification), which represents the degree to which the explanations reflect the GNN's test accuracies. To this end, we acquire a set of GNN models leading to varying accuracies by corrupting portions of labels in the training set from 0\% to 50\% in increment of 5\%. The higher the value of faithfulness, the better the performance.
\end{itemize}


Table~\ref{table:quantitativeeval} summarizes quantitative evaluations for \textsf{PAGE} and XGNN with respect to the \textcolor{black}{accuracy, density, consistency, and faithfulness} when six datasets are used. Table~\ref{table:quantitativeeval}a shows that \textsf{PAGE} consistently provides more accurate explanations when compared to ground truth explanations regardless of datasets. Table~\ref{table:quantitativeeval}b shows that the explanations produced by \textsf{PAGE} tend to exhibit a lower density in most cases, which implies that the explanations are relatively less complex. In Table~\ref{table:quantitativeeval}c, we observe that \textsf{PAGE} provides more stable explanations across different hidden dimension configurations than those of XGNN for all datasets except MNIST-sp (note that such an exception is mainly due to output probabilities of XGNN being near zero for almost all the cases, resulting in a lower standard deviation). From Table~\ref{table:quantitativeeval}d, we observe that \textsf{PAGE} apparently exhibits a positive correlation between the GNN's test accuracies and the output probabilities compared to XGNN, which implies that \textsf{PAGE} is able to successfully account for what the GNN model has learned about its explanation outcome.

To further analyze the faithfulness for \textsf{PAGE} and XGNN, we visualize scatter plots of the output probabilities of explanations versus the GNN's test accuracies in Fig.~\ref{figure:quantfaith}. The intriguing observations are made from Table~\ref{table:quantitativeeval}d and Fig.~\ref{figure:quantfaith}:

\begin{itemize}
\item The different tendency between \textsf{PAGE} and XGNN is apparent on BA-house, BA-grid, Benzene, and MNIST-sp. \textcolor{black}{For Solubility and MUTAG}, the GNN's test accuracy tends to be quite low when labels are corrupted, which thus does not exhibit a clear relationship between the output probability and the test accuracy. This implies that the two datasets are more challenging to be learned in such noisy settings.

\item Overall, the performance difference \textsf{PAGE} and XGNN in terms of the faithfulness becomes more prominent for the synthetic datasets. This is because the ground truth explanation can establish a relationship with the corresponding class more clearly due to less noise of the synthetic datasets.

\item XGNN reveals a negative Kendall's tau coefficient for some datasets. 

\item This is because, when the GNN model is trained with high label corruption ratios, it basically passes through as a random classifier and thus cannot decide whether the explanation generated from XGNN is valid or not. When the GNN model is trained with fewer corrupted data, it now learns the characteristics of the given dataset and will produce low output probabilities if the outcome of XGNN does not contain the complete ground truth prototype. Eventually, this results in a negative correlation between the output probability versus the GNN's test accuracy.

\end{itemize}

\subsubsection{Impact of Multiple Search Sessions in \textsf{PAGE} (RQ3)}

\begin{figure}[t]
\centering
\includegraphics[width=\columnwidth]{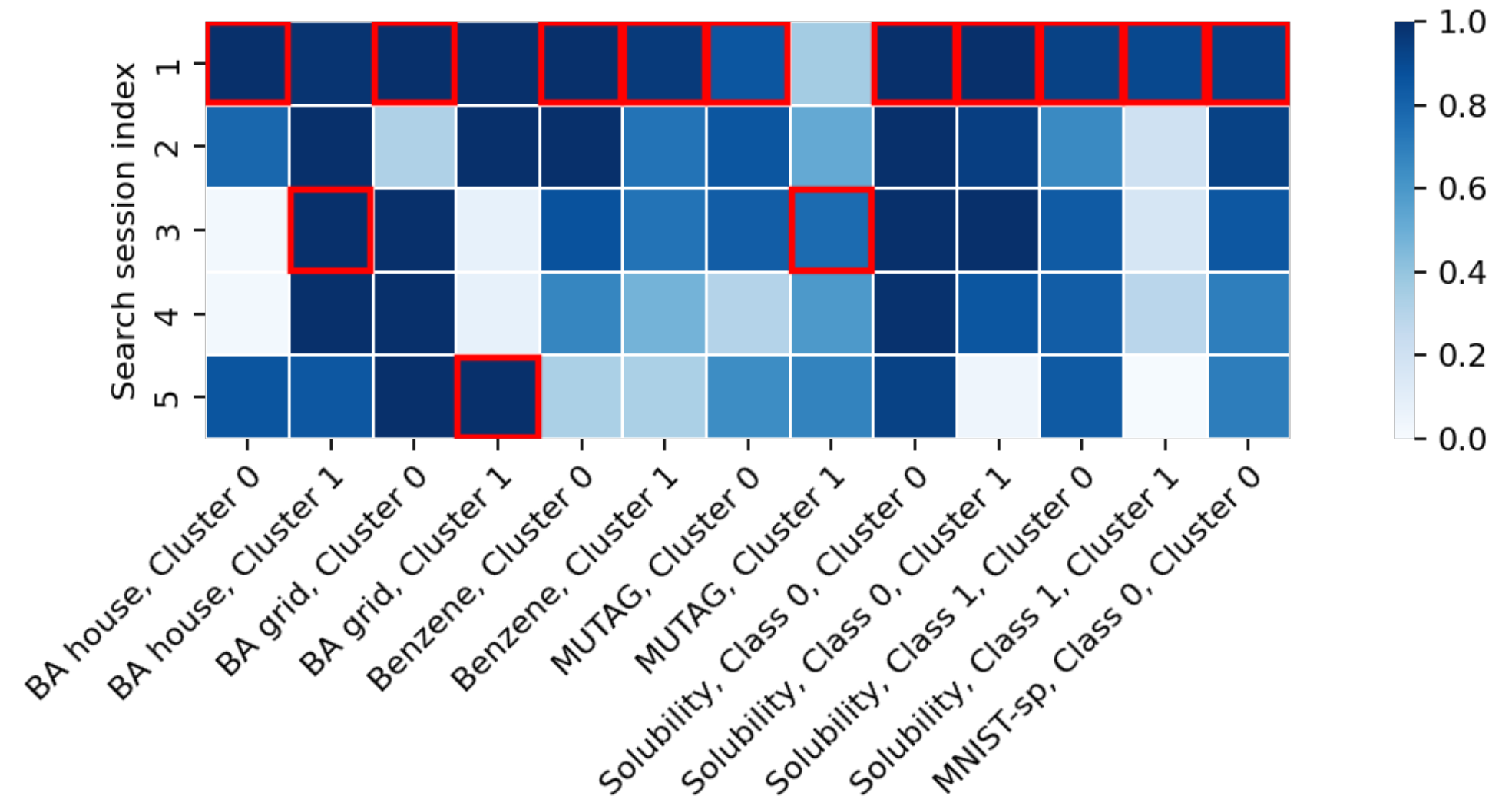}
\caption{Heatmap visualization of the output probability for the discovered subgraph in each search session when \textsf{PAGE} is used for all datasets. The subgraph returned as the final prototype graph is marked with a red box.}
\label{figure:qual0-1}
\end{figure}

Aside from the final prototype graph of \textsf{PAGE}, we are also interested in analyzing the output probability $p_\text{GNN}$ calculated for each search session in order to see when the prototype candidate having the highest $p_\text{GNN}$ is selected over {\it budget} search sessions. Fig.~\ref{figure:qual0-1} visualizes a heatmap of the output probability $p_\text{GNN}$ for the discovered subgraph (i.e., the prototype candidate) in each session for all datasets, being partitioned into two clusters for a certain class, when {\it budget} is given by 5. The prototype candidate chosen as the final prototype graph $g_l^{(c)}$ is highlighted with a red box for each case. We would like to make the following observations:

\begin{itemize}
\item In most cases, each search session tends to return a prototype candidate with various values of $p_\text{GNN}$. For example, in Cluster 1 of MUTAG, the resultant $p_\text{GNN}$'s of the discovered prototype candidates range from the values near 0.4 to the ones near 0.8, which faithfully reflects our intention for the design of the core prototype search module (see Section~\ref{subsubsection:mainproto}).
\item The first search session produces the final explanation (i.e., the final prototype graph $g_l^{(c)}$) for 7 out of 10 cases. This tendency is expected since the first search session starts the iterative subgraph search (Step 2 of Phase 2 in Section~\ref{subsubsection:iterativesubgraphsearch}) alongside the highest matching score in ${\bf Y}_l$.
\item There are a few cases in which the prototype candidate returned by a search session other than the first one is chosen as the final prototype graph, thereby justifying the necessity of multiple search sessions.
\item From these observations, one can see that the number of search sessions, namely {\it budget}, will not be a critical problem in determining the performance of \textsf{PAGE}.
\end{itemize}

\subsubsection{Relation Between PAGE and Instance-Level Explanations (RQ4)} \label{section:RQ4}

Since most studies on GNN explanations focus on \textcolor{black}{instance-level explanations}, we perform an additional analysis by investigating the relationship between our proposed \textsf{PAGE} method and instance-level explanation methods. \textcolor{black}{Specifically, we are interested in evaluating the attribution map of the prototype graph, which indicates its significance towards making the model decisions. To this end, after discovering the prototype graph $g_l^{(c)}$ along with \textsf{PAGE}, we use $g_l^{(c)}$ as the input to an instance-level explanation method to acquire the attribution map (also known as the saliency map).} This evaluation can be interpreted as measuring the agreement between \textsf{PAGE} and instance-level explanation methods.

For a graph $G_i = (\mathcal{V}_i, \mathcal{E}_i, \mathcal{X}_i)$, an instance-level explanation method takes a node $v$ in a graph as input and returns its attribution score, denoted as $\Lambda(v)$, which is normalized across all nodes in the graph. We present the following two quantities that we define in our experiments.
\begin{definition}[Concentration score]
Given a prototype graph $g_l^{(c)}$ and the ground truth explanation $g_T^{(c)}$, we define the {\it concentration score} $\alpha$ as follows:
\begin{equation}
    \alpha = \sum_{v \in \mathcal{V}_{g_T^{(c)}} \cap \mathcal{V}_{g_l^{(c)}}} \Lambda(v),
\end{equation}
\end{definition}
\noindent where $\mathcal{V}_{g_T^{(c)}}$ and $\mathcal{V}_{g_l^{(c)}}$ are the sets of nodes belonging to subgraphs $g_T^{(c)}$ and $g_l^{(c)}$, respectively. The concentration score measures how much the instance-level explanation concentrates on the nodes belonging to the ground truth explanation $g_T^{(c)}$. If none of the nodes in the prototype $g_l^{(c)}$ belongs to $g_T^{(c)}$, then $\alpha$ becomes zero. The higher the value of $\alpha$ lying between 0 and 1, the better the performance.

\begin{figure}[t]
\centering
\begin{subfigure}[b]{\columnwidth}
\centering
   \includegraphics[width=0.9\linewidth]{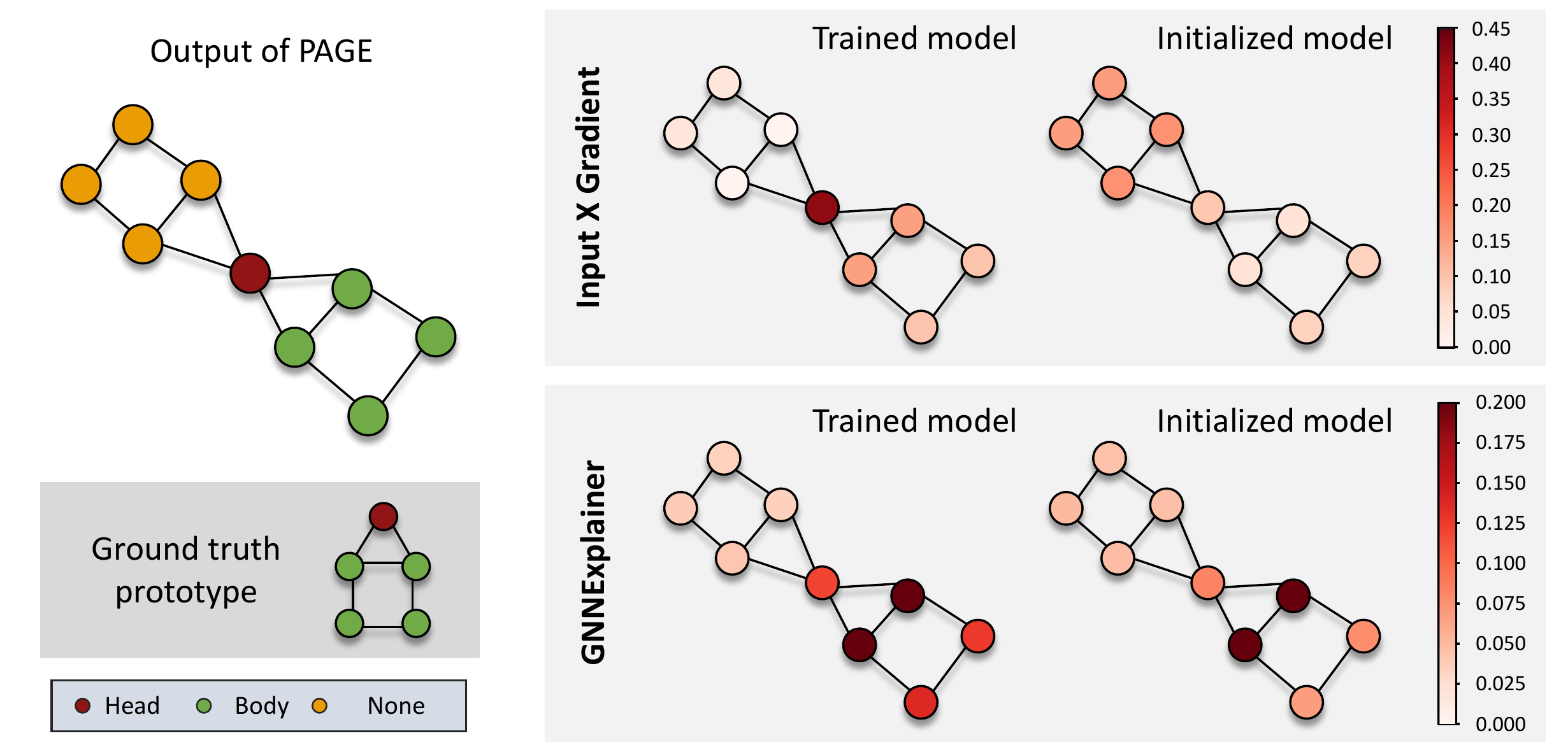}
   \caption{BA-house.}
   \label{figure:rel_ba-house} 
\end{subfigure}

\begin{subfigure}[b]{\columnwidth}
\centering
   \includegraphics[width=0.9\linewidth]{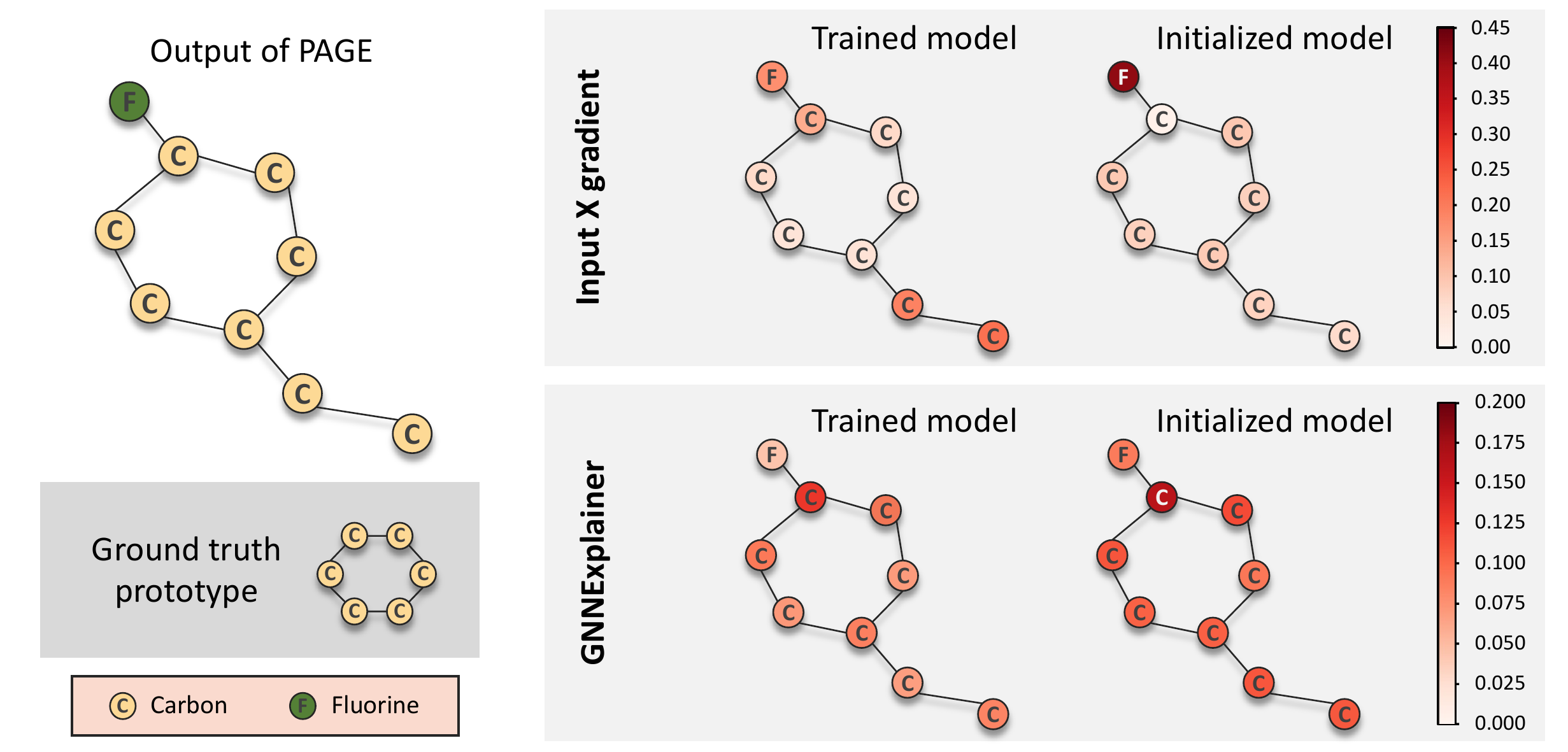}
   \caption{Benzene.}
   \label{figure:rel_benzene}
\end{subfigure}
\caption{Visualization of the attribution score as a heatmap where Input $\times$ Gradient and GNNExplainer are adopted to the output prototype graphs from \textsf{PAGE} for the BA-house and Benzene datasets.}
\label{figure:rel_evaluation}
\end{figure}

\begin{definition}[Relative training gain] Given the concentration score $\alpha$ for a prototype graph $g_l^{(c)}$ and the ground truth explanation $g_T^{(c)}$, the {\it relative training gain} $\beta$ is defined as

\begin{equation}
    \beta = \dfrac{\alpha}{\alpha_0} - 1,
\end{equation}
where $\alpha_0$ denotes the concentration score calculated using an initialized GNN model before training for the same prototype graph $g_l^{(c)}$.
\end{definition}

The relative training gain measures the relative gain of the concentration score after the GNN model has been trained. The higher the value of $\beta$, the better the performance. Note that $\beta$ differs from the faithfulness in that it focuses on the {\em relationship} between model-level and instance-level explanation methods. 

In our experiments, we adopt the following two instance-level explanation methods:
\begin{itemize}
    \item {\bf Input $\times$ Gradient~\cite{shrikumar2016inputgrad}:} This method takes the gradients of the output with respect to the input and multiplies by the input in order to produce attribution maps.
    \item {\bf GNNExplainer~\cite{ying2019gnnexplainer}:} Given an input instance, this method identifies a subgraph structure and a small subset of node features that are most influential to GNN's prediction. GNNExplainer performs an optimization task in the sense of maximizing the mutual information between the prediction and the distribution of possible subgraph structures.
\end{itemize}



\begin{table}[t]
  \centering
  \subfloat[BA-house.]{%
  \begin{tabular}{p{25mm}>{\centering\arraybackslash}m{13mm}>{\centering\arraybackslash}m{13mm}>{\centering\arraybackslash}m{13mm}}
        \toprule
        \textbf{Method} & $\alpha$ & $\beta$ & AUROC \\
        \midrule
        Input $\times$ Gradient  & 0.9128 & 1.5919 & 1.0000\\
        GNNExplainer  & 0.8458 & 0.0477 & 1.0000 \\
        \bottomrule
  \end{tabular}
    }

  \subfloat[Benzene.]{%
  \begin{tabular}{p{25mm}>{\centering\arraybackslash}m{13mm}>{\centering\arraybackslash}m{13mm}>{\centering\arraybackslash}m{13mm}}
        \toprule
        \textbf{Method} & $\alpha$ & $\beta$ & AUROC\\
        \midrule
        Input $\times$ Gradient  & 0.4193 & -0.0592 & 0.0000 \\
        GNNExplainer  & 0.7028 & 0.0170 & 0.8889 \\
        \bottomrule
  \end{tabular}
    }
\caption{Quantitative assessment results of measuring the agreement between \textsf{PAGE} and two instance-level explanation methods for the BA-house and Benzene datasets, along with the accuracy~\cite{sanchez2020gnneval} of the instance-level explanation method methods measured in terms of the AUROC.}\label{table:relationexperiment}
\end{table}

Table~\ref{table:relationexperiment} summarizes quantitative evaluations for measuring the agreement between \textsf{PAGE} and two instance-level explanation methods with respect to the concentration score $\alpha$ and the relative training gain $\beta$ when the BA-house and Benzene datasets are used, along with the performance of each instance-level explanation method by measuring the area under the receiver operating characteristic (AUROC) against the ground truth prototype. To further analyze the relation between \textsf{PAGE} and two instance-level explanation methods, we visualize the attribution score as a heatmap where the two instance-level explanation methods are adopted to the output prototype graphs discovered by \textsf{PAGE} for the two datasets in Fig.~\ref{figure:rel_evaluation}.

We first analyze the quantitative assessment results with respect to $\alpha$ in Table~\ref{table:relationexperiment} and Fig.~\ref{figure:rel_evaluation}.

\begin{itemize}
\item For the BA-house dataset, the values of $\alpha$ achieved by both Input $\times$ Gradient and GNNExplainer are sufficiently high, which indicates that a vast majority of attribution scores are assigned to the nodes belonging to the house-shaped subgraphs. In other words, the performance of \textsf{PAGE} and two instance-level explanation methods is shown to behave closely and consistently with each other.

\item For the Benzene dataset, although the above tendency is observed similarly for GNNExplainer, the value of $\alpha$ tends to diminish for Input $\times$ Gradient. It is known that gradient-based attribution methods may often fail to capture the GNN model's behavior since activation functions such as Rectified Linear Units (ReLUs) have a gradient of zero when they are not activated during the feed-forward process of GNN~\cite{shrikumar2016inputgrad}.
\end{itemize}

We turn to analyzing the quantitative assessment results with respect to $\beta$ in Table~\ref{table:relationexperiment}.

\begin{itemize}
\item One can expect to see a positive value of $\beta$ if the performance of both \textsf{PAGE} and instance-level explanation methods behaves closely and consistently. 

\item However, a negative value of $\beta$ is observed for the case of Input $\times$ Gradient on Benzene, which is in contrast to the case of GNNExplainer producing positive values of $\beta$ for the two datasets.

\item The higher agreement between $\textsf{PAGE}$ and GNNExplainer on Benzene is due to the fact that GNNExplainer was designed for generating explanations on graphs and GNN models, thus reliably producing the resulting attributions. 
\end{itemize}

The above tendency is also found with respect to the AUROC in Table~\ref{table:relationexperiment}, where Input $\times$ Gradient returns zero as AUROC compared to the case of GNNExplainer returning the AUROC of 0.8889. Note that the results from other three datasets showed a tendency similar to those reported in Section~\ref{section:RQ4}.

\subsubsection{Robustness to the Number of Available Input Graphs (RQ5)}

We now perform a qualitative evaluation of \textsf{PAGE} by visualizing the resultant prototype graphs in a more difficult setting where each dataset is composed of only a few available graphs $\tilde{\mathcal{G}}$ from the training set as input of \textsf{PAGE}. This often occurs in real environments since users and organizations aiming at designing explanation methods may have limited access to the input data. To emulate such a scenario, we create the subset of cardinality 10 (i.e., $|\tilde{\mathcal{G}}| = 10$) via random sampling from the training set. In other words, the subset $\tilde{\mathcal{G}}$ is used for running \textsf{PAGE} while the GNN model has been trained using the whole training set from $\mathcal{G}$. Due to the fact that clustering of graph-level embeddings in Phase 2 of \textsf{PAGE} is not possible alongside only a few graphs in $\tilde{\mathcal{G}}$, we simplify \textsf{PAGE} and perform Phase 2 only after selecting $k$ graphs from $\tilde{\mathcal{G}}$ instead of the $k$NNs.

\begin{figure}[t]
\centering
\begin{subfigure}[b]{\columnwidth}
\centering
   \includegraphics[width=\textwidth]{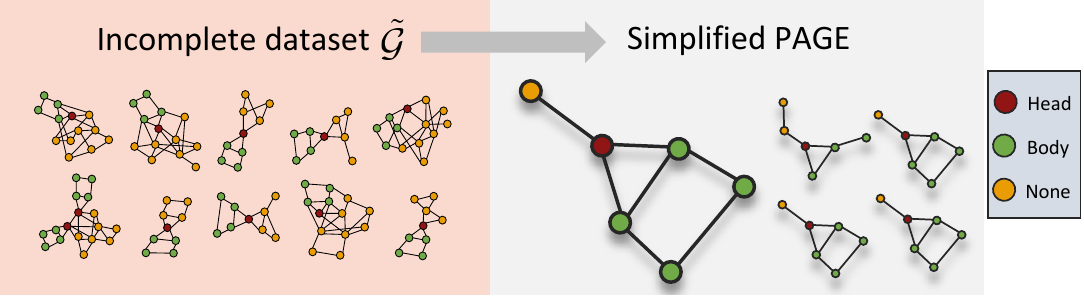}
   \caption{BA-house.}
   \label{figure:incomplete_ba-house} 
\end{subfigure}
\begin{subfigure}[b]{\columnwidth}
\centering
   \includegraphics[width=\textwidth]{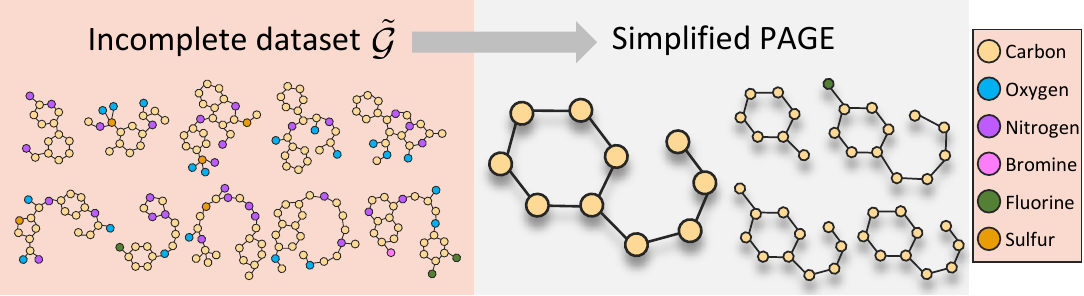}
   \caption{Benzene.}
   \label{figure:incomplete_benzene}
\end{subfigure}
\caption{Qualitative analysis of \textsf{PAGE} in an incomplete dataset setting, where each node is colored differently according to the types of nodes.}
\label{figure:incomplete_evaluation}
\end{figure}

Fig.~\ref{figure:incomplete_evaluation} visualizes the prototype graphs discovered by simplified \textsf{PAGE} for five independent trials when the subset $\tilde{\mathcal{G}}$ of the BA-house and Benzene datasets is used for $k=3$. In the case of BA-house in Fig.~\ref{figure:incomplete_ba-house}, we observe that the prototype contains an entire single house-shaped subgraph for four out of five trials, indicating that the core prototype search module is capable of producing ground truth explanations even in the incomplete dataset setting. However, due to the absence of clustering, the prototype graph containing double house-shaped subgraphs is not retrieved. Therefore, the simplified \textsf{PAGE} does not produce precise explanations but resorts to simpler explanations. In the case of Benzene in Fig.~\ref{figure:incomplete_benzene}, we observe that the carbon ring structure is included for all five trials; the number of atoms other than the ground truth explanation varies over different trials due to the random selection of graph subsets $\tilde{\mathcal{G}}$. Note that the results from other three datasets also follow similar trends although not shown in the manuscript.

\subsubsection{Effectiveness of the Prototype Scoring Function (RQ6)}
We validate the effectiveness of our prototype scoring function $s$ in Definition~\ref{definition:scoringfunction} in terms of the computation efficiency in comparison to a na\"ive alternative. Instead of acquiring a set of node-level embedding vectors through a certain GNN model, we synthetically generate and feed them into the function $s$, which is sufficient to analyze the effectiveness of $s$. Specifically, we first generate each Gaussian random vector ${\bf w}_i$ with zero mean and covariance matrix ${\bf I}_b$, i.e., ${\bf w}_i\sim \mathcal{N}({\bf 0}_b, {\bf I}_b)$, where ${\bf 0}_b$ and ${\bf I}_b$ denote the zero vector and the identity matrix, respectively, of size $b$. Then, the vectors ${\bf w}_i$'s pass through the ReLU activation function ${\bf v}_i=\text{ReLU}({\bf w}_i)$, which corresponds to a synthetic node-level embedding vector. 

In our experiments, we generate $10^3$ 3-tuples of vectors, which are used as the input of $s({\bf v}_1,{\bf v}_2,{\bf v}_3)$. We compare $s$ with the following alternative prototype scoring functions: $s_{AM}({\bf v}_1,{\bf v}_2,{\bf v}_3)=({\bf v}_1^T{\bf v}_2 + {\bf v}_2^T{\bf v}_3 + {\bf v}_3^T{\bf v}_1)/3$ and $s_{GM}({\bf v}_1,{\bf v}_2,{\bf v}_3)=({\bf v}_1^T{\bf v}_2 \times {\bf v}_2^T{\bf v}_3 \times {\bf v}_3^T{\bf v}_1)^{1/3}$, which return the arithmetic mean and the geometric mean, respectively, of all possible inner products between pairs of embedding vectors.

\begin{table}[t]
\centering
  \begin{tabular}{lccc}
    \toprule
    Prototype & \multirow{2}{*}{$s$} & \multirow{2}{*}{$s_{\text{AM}}'$} & \multirow{2}{*}{$s_{\text{GM}}'$} \\
    scoring function &  &  &  \\
    \midrule
    Time ($\mu s$)  & 14.42 $\pm$ 12.65 & 38.34 $\pm$ 17.81  & 31.39 $\pm$ 17.81 \\
    \bottomrule
  \end{tabular}
\caption{Comparison of different prototype scoring functions in terms of the runtime complexity in microseconds (mean $\pm$ standard deviation).}
\label{table:comput_evaluation}
\end{table}

\begin{figure}[t]
\centering
\begin{subfigure}{.5\columnwidth}
  \centering
  \includegraphics[width=0.9\linewidth]{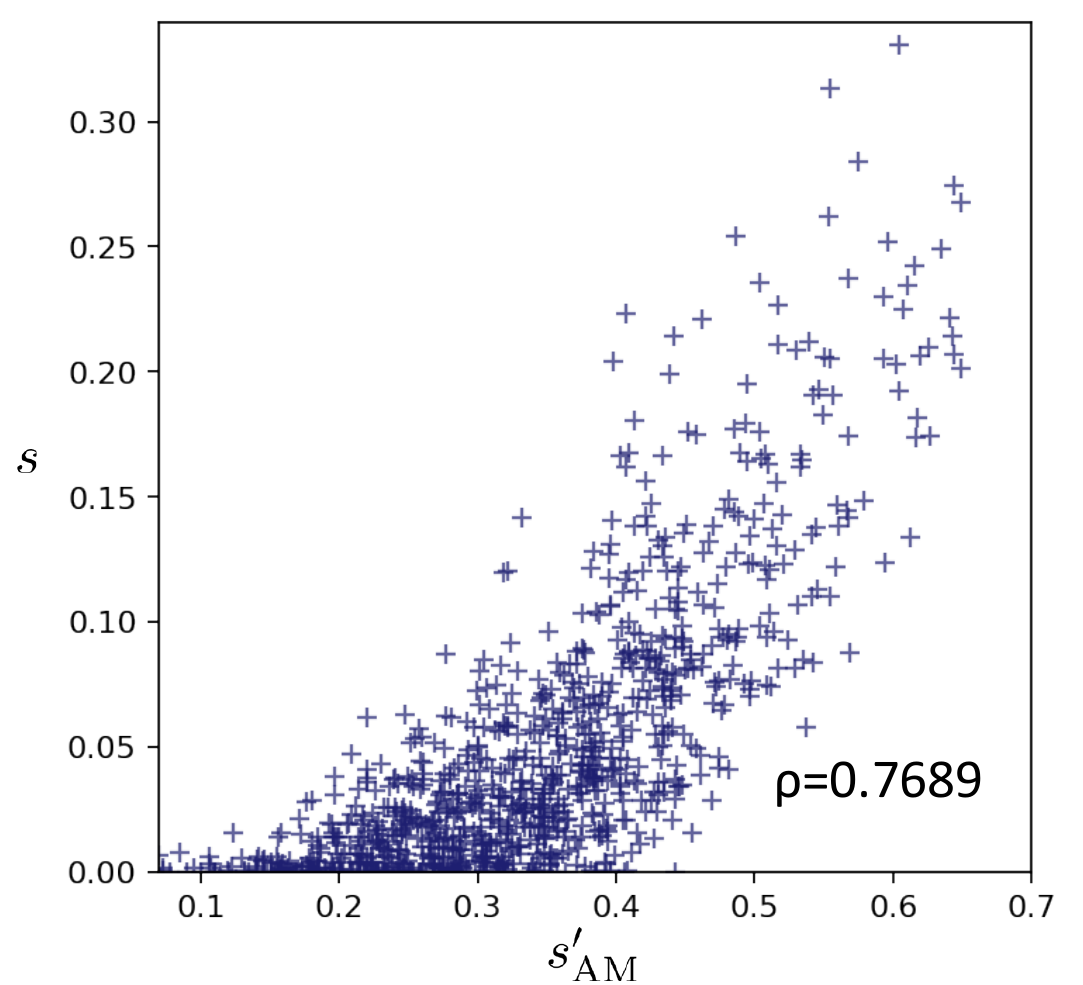}
  \caption{$s_{\text{AM}}'$ versus $s$}
  \label{fig:comput_a}
\end{subfigure}%
\begin{subfigure}{.5\columnwidth}
  \centering
  \includegraphics[width=0.9\linewidth]{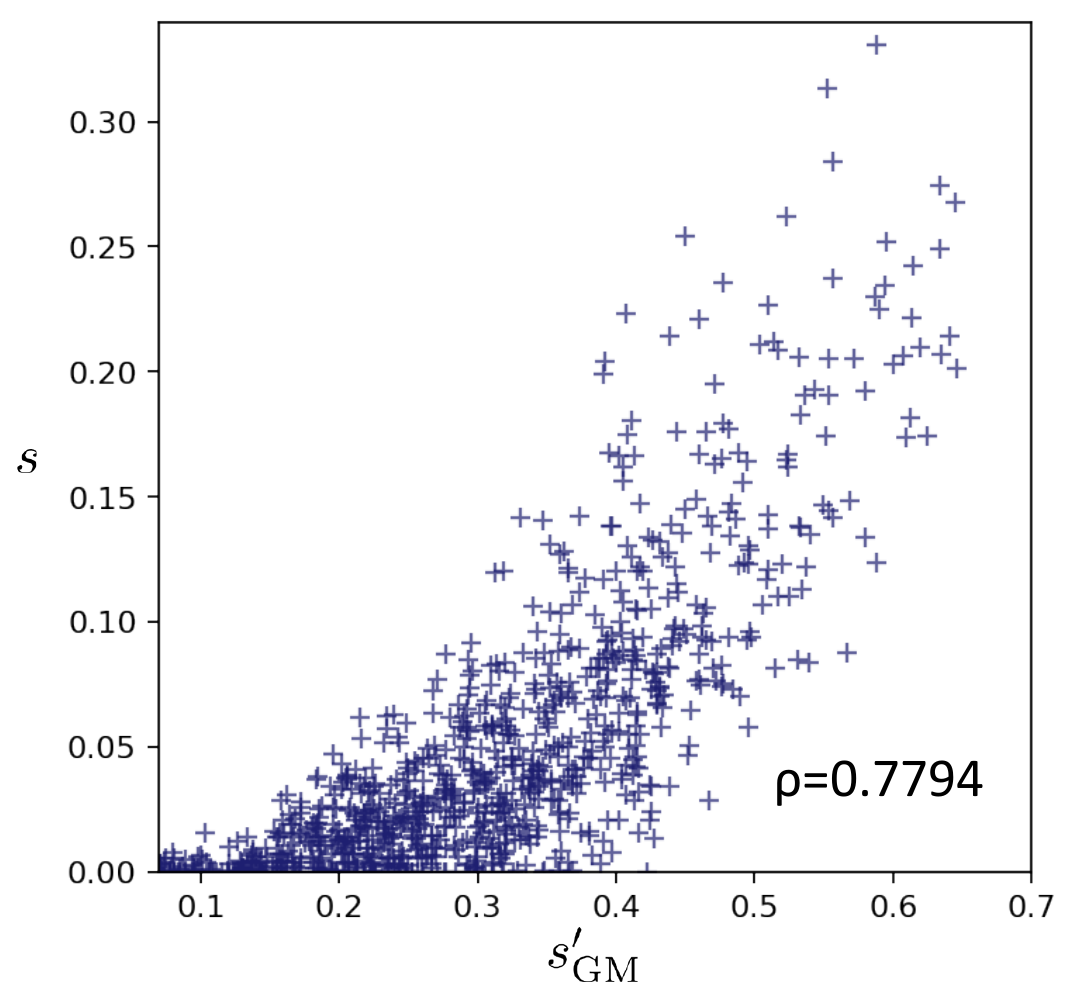}
  \caption{$s_{\text{GM}}'$ versus $s$}
  \label{fig:comput_b}
\end{subfigure}
\caption{Visualization of the results of different prototype scoring functions along with the correlation coefficient $\rho$.}
\label{fig:comput_evaluation}
\end{figure}

We empirically show the average runtime complexities of $s$, $s_{AM}$, and $s_{GM}$ in Table~\ref{table:comput_evaluation}. This experiment reveals that $s_{AM}$ and $s_{GM}$ are more computationally expensive than $s$, which is consistent with our analysis in Remark~\ref{remark:scoringfunction}. Additionally, Fig.~\ref{fig:comput_evaluation} visualizes scatter plots of the proposed $s$ versus each of the na\"{i}ve alternative, i.e., either $s_{AM}$ or $s_{GM}$. From this figure, we observe a high correlation between two, where the Pearson correlation coefficient $\rho$ is more than 0.75 for both cases. This demonstrates that our prototype scoring function $s$ is a good approximation of the calculation of all possible inner products between embedding pairs.

\subsubsection{Efficiency of \textsf{PAGE} (RQ7)}
We validate the superiority of \textsf{PAGE} in terms of the efficiency. To this end, we first theoretically analyze its computational complexity. In Phase 1, the complexity of GMM is $\mathcal{O}(nkb^3)$~\cite{pinto2015gmmcomplexity} and the $k$NN selection involves sorting for each cluster, which results in $\mathcal{O}(nk(b^3 + \log n))$, where $b$ is the dimension of node-level embedding vectors. In Phase 2, since computing ${\bf Y}_l$ and running each search session are highly parallelizable, the computation depends basically on the pre-defined maximum allowed number of iterations during the iterative subgraph search, $N_{\text{max}}$, and the number of non-zero elements in ${\bf Y}_l$, $N_{\text{nnz}}$. Thus, the resulting complexity for Phase 2 of \textsf{PAGE} is $\mathcal{O}(N_{\text{max}} N_{\text{nnz}} \log N_{\text{nnz}}).^6$ Therefore, the total computational complexity of \textsf{PAGE} is given by $\mathcal{O}(nk(b^3 + \log n) + N_{\text{max}} N_{\text{nnz}} \log N_{\text{nnz}})$.

Additionally, we conduct experiments using two datasets, BA-house and Benzene, to measure the runtime complexities of \text{PAGE} and XGNN. Table~\ref{table:runtimecomparison} the runtime complexity (mean $\pm$ standard deviation) of each model-level GNN explanation method for 10 independent trials. The results demonstrate the efficiency of \textsf{PAGE} over XGNN regardless of the datasets.

\begin{table}[t]
\centering
\resizebox{0.65\columnwidth}{!}{%
  \begin{tabular}{lcc}
    \toprule
    Dataset & \textsf{PAGE} & XGNN \\
    \midrule
    BA-house  & 1.15 ± 0.01 $s$ & 3.14 ± 0.02 $s$ \\
    Benzene  & 12.10 $\pm$ 0.02 $s$ & 13.50 $\pm$ 0.14 $s$ \\
    \bottomrule
  \end{tabular}
}
\caption{Comparison of runtime complexities between \textsf{PAGE} and XGNN on the BA-house and Benzene datasets.}
\label{table:runtimecomparison}
\end{table}

\color{black}

\begin{figure}[t]
     \centering
     \begin{subfigure}[b]{0.44\textwidth}
         \centering
         \includegraphics[width=\textwidth]{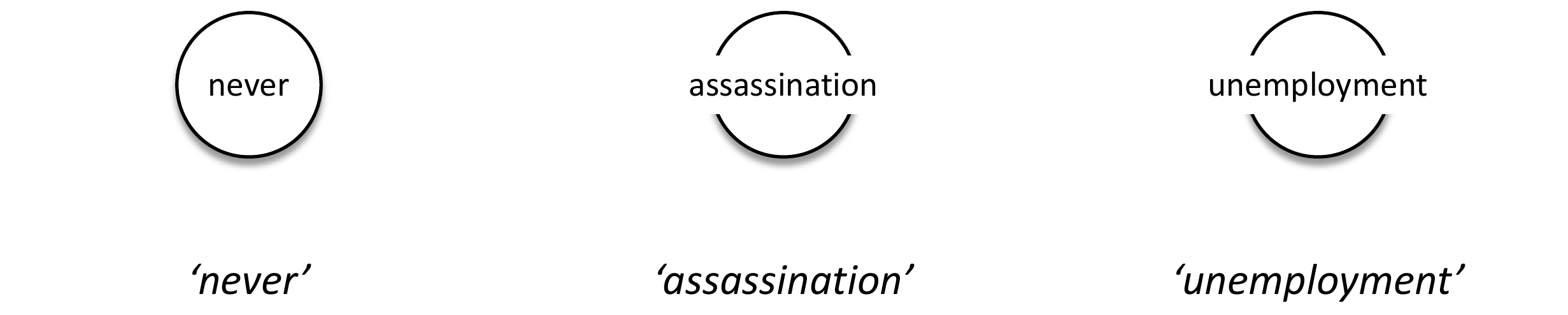}
         \caption{Class 0, Cluster 0.}
         \label{fig:Graph-SST2_result1}
     \end{subfigure}
     \hfill
     \begin{subfigure}[b]{0.44\textwidth}
         \centering
         \includegraphics[width=\textwidth]{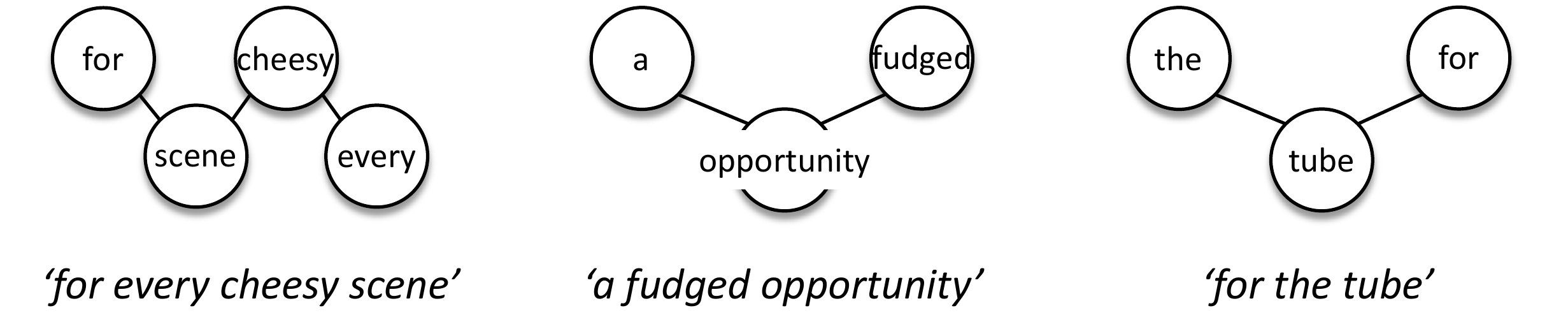}
         \caption{Class 0, Cluster 1.}
         \label{fig:Graph-SST2_result2}
     \end{subfigure}\\
     \begin{subfigure}[b]{0.44\textwidth}
         \centering
         \includegraphics[width=\textwidth]{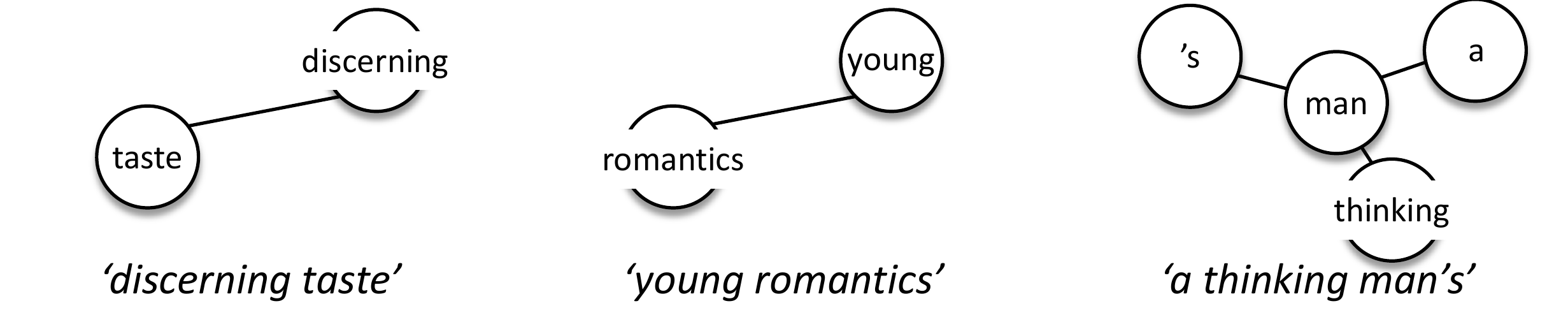}
         \caption{Class 1, Cluster 0.}
         \label{fig:Graph-SST2_result3}
     \end{subfigure}
     \hfill
     \begin{subfigure}[b]{0.44\textwidth}
         \centering
         \includegraphics[width=\textwidth]{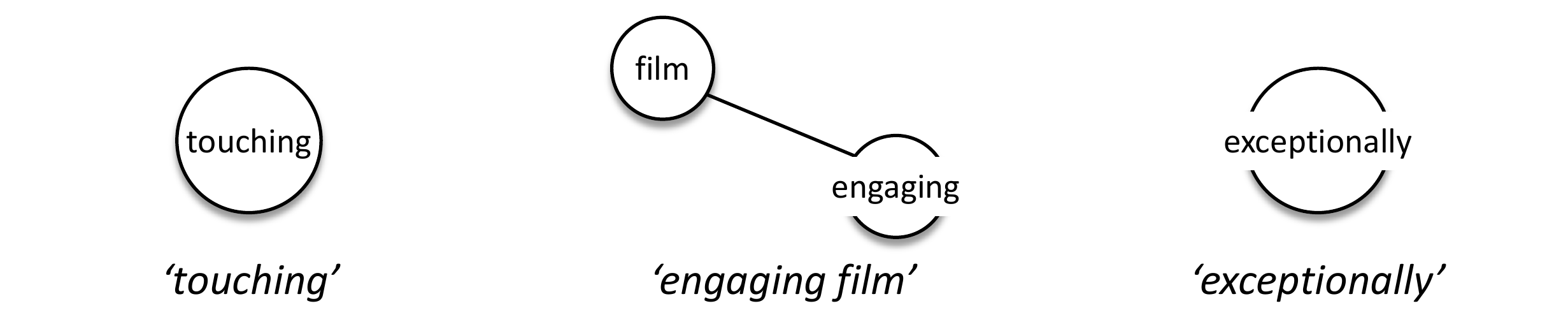}
         \caption{Class 1, Cluster 1.}
         \label{fig:Graph-SST2_result4}
     \end{subfigure}
        \caption{\textcolor{black}{Explanation results of \textsf{PAGE} on the Graph-SST2 dataset.}}
        \label{fig:Graph-SST2_results}
\end{figure}
\vspace{0.25in}

\subsubsection{Performance of \textsf{PAGE} on the Graph-SST2 Dataset (RQ8)}

We evaluate the performance of \textsf{PAGE} on Graph-SST2~\cite{yuan2020xaisurvey}, the dataset without explicit ground truth explanations. Graph-SST2 is a binary graph classification dataset processed from the SST2 dataset~\cite{socher2013sst2}. Sentences from the SST2 dataset are transformed into graphs by representing words as nodes and constructing edges according to the relationships between words. 786-dimensional word embedding vectors are used as node features.

Since Graph-SST2 is a sentiment classification dataset, it does not have a clear {\it ground truth} explanation that commonly expresses itself across instances, posing a critical challenge to run \textsf{PAGE}. In other words, it is technically difficult to discover ground truth explanations (prototypes) representing the whole class. The fact that the edges represent the lexical collocations makes the task more problematic, since the lexical combinations do not have an explicit connection to the classes. 

Despite such a critical limitation, we run a part of \textsf{PAGE} on Graph-SST2 to see what input graphs are selected as the most representative ones. More specifically, we perform Phase 1 on Graph-SST2 as a simplified version of \textsf{PAGE}. Fig.~\ref{fig:Graph-SST2_results} visualizes the results of \textsf{PAGE} by selecting three graphs for each class/cluster. 
From the figure, we would like to make the following observations.
\begin{itemize}
    \item As discussed earlier, extracting a common subgraph pattern is not possible. This is because 1) there are cases where node edge exists and 2) additional criteria are necessary to determine whether a node matches another node to form a tuple. 
    \item The selected input sentences tend to be very short in length. This can be naturally interpreted as a result of Phase 1 in \textsf{PAGE} that attempts to discover a common pattern among examples in a cluster. When a sentence becomes longer, it will be more uniquely positioned in the belonging cluster as it is likely to include rarely used words, thus yielding a unique graph structure. Although \textsf{PAGE} cannot provide explicit prototype graphs in contrast to other datasets classified as chemical/molecular datasets, it can still offer a low-resolution concept of each class representing the sentiment of phrases, which is another merit of \textsf{PAGE} behind.
\end{itemize}

\color{black}

\section{Conclusions
and Outlook} \label{section:conclusion}

We investigated a largely underexplored yet important problem of explaining the decisions of GNNs at the model level. To tackle this practical challenge, we introduced a novel explanation method that offers model-level explanations of GNNs for the graph classification task by discovering prototypes from both node-level and graph-level embeddings. Specifically, we designed \textsf{PAGE}, an effective two-phase model-level GNN explanation method; after performing clustering and selection of graph-level embeddings in Phase 1, we perform prototype discovery in Phase 2, which is partitioned into the initialization, iterative subgraph search, and prototype retrieval steps. Furthermore, we theoretically validated the usage of clustering of graph-level embeddings in \textsf{PAGE}. Using two synthetic and three real-world datasets, we comprehensively demonstrated that \textsf{PAGE} is superior to the state-of-the-art model-level GNN explanation method both qualitatively and quantitatively. Additionally, through systematic evaluations, we proved the effectiveness of \textsf{PAGE} in terms of 1) the impact of multiple search sessions in the core prototype search module, 2) the relation to instance-level explanation methods via quantitative assessment, 3) the robustness to a difficult and challenging situation where only a few graphs are available as input of \textsf{PAGE}, and 4) the computational efficiency of the proposed prototype scoring function.

The main limitation of our work mainly lies in the assumption that the number of ground truth explanations per class is available, under which we run Phase 1 in \textsf{PAGE} to discover multiple clusters on the graph-level embedding space. In Phase 1, one can use various performance metrics used for the quantitative analysis to find the optimal number of clusters. As an outlook on future research, estimating the number of ground truth explanations per class before running explanation methods is an intriguing and important research topic from the fact that Phase 1 of \textsf{PAGE} needs the number of ground truth explanations per class for clustering, which is however not known a priori. Additional avenues of future research include the design of model-level GNN explanation methods for the node-level or edge-level classification task.

\appendices

\ifCLASSOPTIONcompsoc
  \section*{Acknowledgments}
\else
  \section*{Acknowledgment}
\fi

This research was supported by the National Research Foundation of Korea (NRF) grant funded by the Korea government (MSIT) (No. 2021R1A2C3004345, No. RS-2023-00220762). The material in this paper was presented in part at the AAAI Conference on Artificial Intelligence, Virtual Event, February/March 2022~\cite{Shin2022EarlyPAGE}.

\ifCLASSOPTIONcaptionsoff
  \newpage
\fi

\bibliographystyle{IEEEtran}
\bibliography{bibliography.bib}


\end{document}